\documentclass[10pt,journal,compsoc]{IEEEtran}
\usepackage{graphicx}
\usepackage{amsmath}
\usepackage{color}
\usepackage{multirow}
\usepackage{amssymb}
\usepackage[table,xcdraw]{xcolor}

\ifCLASSOPTIONcompsoc
  \usepackage[nocompress]{cite}
\else
  \usepackage{cite}
\fi
\ifCLASSINFOpdf
\else
\fi
\begin{document}
\title{Dynamic MDETR: A Dynamic Multimodal Transformer Decoder for Visual Grounding}

\author{Fengyuan Shi, Ruopeng Gao, Weilin Huang, 
        Limin Wang,~\IEEEmembership{Member,~IEEE}% <-this % stops a space
\IEEEcompsocitemizethanks{
\IEEEcompsocthanksitem F. Shi, R. Gao, and L. Wang are with State Key Laboratory for Novel Software Technology, Nanjing University, China. (email: fengyuanshi1999@gmail.com, ruopenggao@gmail.com, lmwang@nju.edu.cn) (Corresponding author: Limin Wang)
\IEEEcompsocthanksitem W. Huang is with Alibaba Group, China. (email: weilin.hwl@alibaba-inc.com)% <-this % stops an unwanted space
% \IEEEcompsocthanksitem Corresponding authors: L. Wang. 
}
}

\IEEEtitleabstractindextext{%
\begin{abstract}
Multimodal transformer exhibits high capacity and flexibility to align image and text for visual grounding. However, the existing encoder-only grounding framework (e.g., TransVG) suffers from heavy computation due to the self-attention operation with quadratic time complexity.
To address this issue, we present a new multimodal transformer architecture, coined as \textbf{Dynamic} \textbf{M}utilmodal \textbf{DETR} (Dynamic MDETR), by decoupling the whole grounding process into encoding and decoding phases. The key observation is that there exists high spatial redundancy in images. 
Thus, we devise a new dynamic multimodal transformer decoder by exploiting this sparsity prior to speed up the visual grounding process. Specifically, our dynamic decoder is composed of a 2D adaptive sampling module and a text guided decoding module. The sampling module aims to select these informative patches by predicting the offsets with respect to a reference point, while the decoding module works for extracting the grounded object information by performing cross attention between image features and text features. These two modules are stacked alternatively to gradually bridge the modality gap and iteratively refine the reference point of grounded object, eventually realizing the objective of visual grounding.
Extensive experiments on five benchmarks demonstrate that our proposed Dynamic MDETR achieves competitive trade-offs between computation and accuracy. Notably, using only 9\% feature points in the decoder, we can reduce $\sim$44\% GFLOPs of the multimodal transformer, but still get higher accuracy than the encoder-only counterpart. With the same number of encoder layers as TransVG, our Dynamic MDETR (ResNet-50) outperforms TransVG (ResNet-101) but only brings marginal extra computational cost relative to TransVG.
In addition, to verify its generalization ability and scale up our Dynamic MDETR, we build the first one-stage CLIP empowered visual grounding framework, and achieve the state-of-the-art performance on these benchmarks.
\end{abstract}

}

\maketitle

\IEEEdisplaynontitleabstractindextext

% \fi
\IEEEpeerreviewmaketitle

\IEEEraisesectionheading{\section{Introduction}\label{sec:introduction}}

\IEEEPARstart{V}{isual} grounding~\cite{mao2016generation,hu2016natural} aims to localize the object referred by the given expression in an image. Using free-form text descriptions as queries breaks the limitation of restricted categories in conventional object detection task and facilitates to strong generality and usability. Visual grounding is an important multimodal reasoning technology, bridging visual perception and language expression, and has attracted much attention from academia and industry in both computer vision and natural language processing in recent years.

Existing methods for visual grounding can be divided into two categories: (1) two-stage methods~\cite{mao2016generation,hu2016natural,hu2017modeling,yu2018mattnet,wang2019neighbourhood,liu2020learning} and (2) one-stage methods~\cite{yang2019fast,liao2020real,yang2020improving,huang2021look}. Two-stage methods first generate proposals via the off-the-shelf detectors~\cite{ren2015faster, he2017mask} and calculate the similarities between the expression and proposals, then select the best one as the final prediction. One-stage methods usually first fuse language feature into the image feature map, and then directly predict the bounding box in the dense grids with pre-defined anchors~\cite{redmon2018yolov3}. Two-stage methods are capped by the quality of the region candidates produced in the first stage, and cannot be trained in an end-to-end fashion, leading to sub-optimal results~\cite{yang2019fast}. One stage methods can accelerate the inference speed by removing the time-consuming object proposal generation stage. But independent prediction on each point in the grid fails to capture object relations for visual grounding. Moreover, some simple multi-modal fusion operations like concatenation cannot capture cross-modal interactions well. Recently, Transformer~\cite{vaswani2017attention} has achieved significant success in computer vision tasks, including image classification~\cite{dosovitskiy2020image,liu2021swin}, object detection~\cite{carion2020end,zhu2020deformable,gao2022adamixer}, and semantic segmentation~\cite{zheng2021rethinking,xie2021segformer}. There are also some works applying Transformer into the visual grounding task~\cite{deng2021transvg,ye2022shifting,yang2022improving,deng2022transvg++}. For example, TransVG~\cite{deng2021transvg} introduces a multimodal transformer framework to establish intra-modality and inter-modality correspondences with the self-attention mechanism and achieves the state-of-the-art performance. 

\begin{figure*}
    \centering
    \includegraphics[width=0.85\linewidth]{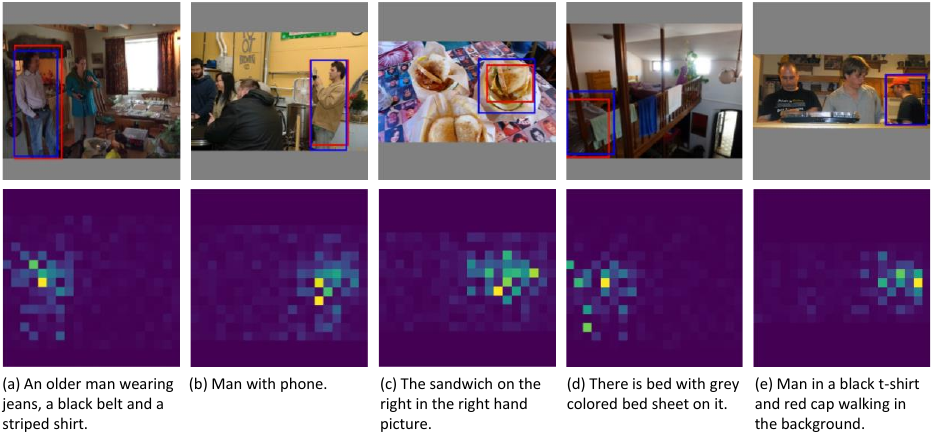}
    \caption{Examples illustrating spatial redundancy in visual grounding task. The top pictures are the padded image as model inputs (640 $\times$ 640). The red box is the ground truth and the blue box is the predicted box by TransVG~\cite{deng2021transvg}. The bottom pictures show the attention map of [REG] token on visual tokens in the last multimodal encoder layer of TransVG. Only a small number of visual tokens have high attention value, and attentions on most of visual tokens in the image is almost zero, showing the severe spatial redundancy in the image for visual grounding task.}
    \label{fig:fig_1}
\end{figure*}

Although multimodal transformer is a simple and effective architecture for visual grounding, it suffers from heavy computation due to the standard attention operation across image and text with quadratic time complexity. In fact, not all positions in the image contribute to the final prediction~\cite{xie2020spatially,rao2021dynamicvit}, such as background area without any object. In this sense, there exists spatial redundancy in the image, which is more severe in visual grounding task, as there not only exists background area without any objects, but also objects unrelated to the target object. For example, Fig.\ref{fig:fig_1} illustrates the ground truth box (red box) and predicted box (blue box), and the attention map of [REG] token on visual tokens in the last multimodal encoder layer of TransVG~\cite{deng2021transvg}. From the attention map, we can find that the attention mainly focuses on the area of the predicted box, and the attention on the other areas of the image is near zero. This result shows that only a small number of visual tokens contribute to the final prediction and there exists serve spatial redundancy in the visual grounding task. 

Some works in single-modality tasks attempt to reduce spatial redundancy in the encoder-only structures by token pruning, e.g., predicting binary mask~\cite{xie2020spatially} on feature map, or selecting the top-k tokens (or above the threshold) based on the informativeness score~\cite{rao2021dynamicvit,pan2021ia}. However, the training of these models is still computationally expensive as all tokens participate in the calculation during training. Meanwhile, it is unclear how to adapt them to the multi-modal tasks as we need to consider multi-modal information correlation in this spatial redundancy reduction process.

To address the aforementioned problem, following the standard DETR framework~\cite{carion2020end} in object detection, we introduce a new multimodal transformer architecture for visual grounding, termed as \textbf{Dynamic} \textbf{M}ultimodal \textbf{DETR} (Dynamic MDETR). Our Dynamic MDETR generally follows the DETR framework and decouples visual grounding task into an encoder-decoder pipeline. Our Dynamic MDETR is based on a relatively shallow encoder for cross-modal feature fusion and alignment, and a dynamic decoder for efficient text-guided visual localization. 
Our core contribution is a multimodal transformer decoder, which eliminates the decoding complexity problem of depending on the image input size. {\em One key design} is the language-guided spatial adaptive sampling in the 2D geometric space. With this sampling module, the dynamic multimodal transformer decoder can adaptively sample a small number of informative visual tokens in the 2D space for subsequent multimodal decoding. This fixed number of sampled tokens would reduce the computational cost to be of constant complexity independent of the image input size. In addition, this adaptive sampling could greatly reduce the influence of these irrelevant tokens on the final bounding box regression, and result in higher performance on the standard benchmarks. Moreover, sampling in 2D geometric space helps to learn positional and geometric information effectively and can lead to better detection performance. {\em The other key design} is text guided decoding, which decodes the sampled visual features into the grounding location of targets under the guidance of text. Using language as queries in our text guided decoding injects strong semantic information about the target object for accurate localization. In this encoder-decoder design, our 2D sampling module simply samples visual features from the same image feature map in each decoder layer without updating the visual features, and our text guided decoding transforms the sampled features for visual grounding. In this sense, this decoding process only focuses on regressing the grounding location of targets and is not responsible for performing multi-modal feature extraction and alignment. Decoupling the whole pipeline into encoder and decoder would enable each module to focus on its own goal: encoder for multimodal feature extraction and fusion, while decoder for efficient sampling and target grounding. On the contrary, adding sampling technique in encoder-only architecture is another possible solution to reduce computational cost, but requires the module to jointly perform feature extraction and token sampling, which would increase the modeling complexity and training difficulty, leading to inferior performance as shown in experiments.

Specifically, in our visual grounding framework of Dynamic MDETR, visual features and language features are first extracted by the respective backbones, and then concatenated into one sequence. The sequence is fed into a multimodal transformer encoder composed of several vanilla attention layers for cross-modal feature fusion and alignment. After this, in the dynamic multimodal transformer decoder, we first inject the language information to the learnable sampling query. Based on this sampling query, our dynamic decoder generates offsets relative to the reference point for feature sampling in the 2D feature map. The sampled features are decoded with the standard transformer under the guidance of language queries. Finally, a simple FFN prediction head directly regresses 4-dim coordinates of the bounding box for the target object. 

The experimental results demonstrate that Dynamic MDETR achieves competitive trade-offs between computation and accuracy. By replacing the computationally expensive transformer encoders with our proposed efficient dynamic multimodal transformer decoders, which use only 9\% feature points as input, we can reduce $\sim$44\% GLOPs of the multimodal transformer encoder but still get higher accuracy on almost all datasets, such as the rise of 1.54 points on the validation set of RefCOCOg-umd~\cite{nagaraja2016modeling}.  
We further investigate model scalability by using jointly trained vision and language model instead of independently trained ones as our feature encoders. With CLIP~\cite{radford2021learning} as backbones, Dynamic MDETR achieves state-of-the-art performance, which also demonstrates the strong localization ability of CLIP in the one-stage framework. To our best knowledge, this is the first work to verify the great performance of CLIP representations in the one-stage visual grounding task. We believe the reason is that an efficient decoder head is necessary to adapt the powerful pre-trained multi-modal encoder in visual grounding task. We hope our work can draw attention to research on efficient multimodal transformer-like models, especially for language guided localization and dense prediction tasks with high-resolution images as input. 

Our contributions are summarized as follows:
\begin{itemize}
    \item 
        We propose an efficient multimodal transformer framework for visual grounding, termed as Dynamic MDETR, with a relatively shallow encoder for cross-modal feature fusion and alignment, and a dynamic decoder for efficient text-guided visual localization.
    \item 
        With 2D adaptive sampling, our proposed dynamic decoder samples a small number of visual features in the 2D space, and reduces the cost to be of constant complexity independent of the image input size.
    \item 
        We conduct exhaustive ablation studies on the encoder-decoder design for in-depth analysis of the encoding-decoding framework in visual grounding task.
    \item 
        The experiments demonstrate that Dynamic MDETR achieves competitive trade-offs between computation and accuracy. With only 9\% feature points as input, we can reduce ~44\% GLOPs of the multimodal transformer encoder but achieve higher accuracy on almost all datasets.
    \item 
        We further investigate CLIP backbone and present the first CLIP empowered visual grounding pipeline. The significant performance improvement shows the generalization ability and scalability of our Dynamic MDETR framework.
\end{itemize}

\begin{figure*}
\centering
\includegraphics[width=\linewidth]{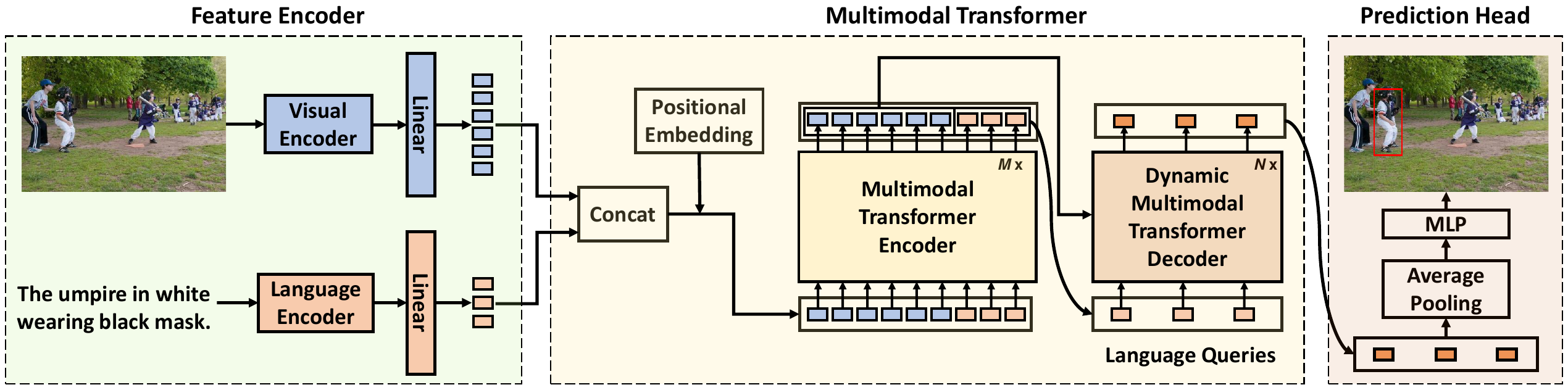}
\caption{The framework of Dynamic MDETR. Dynamic MDETR consists of four main components, including feature encoder, multimodal transformer encoder, dynamic multimodal transformer decoder and prediction head. Specifically, the input image and referring expression are fed into the visual encoder and language encoder respectively. After getting visual features and language features, we concatenate them into a sequence. Adding a learnable positional embedding, we input the sequence into the multimodal transformer encoder for multimodal fusion. In the dynamic multimodal transformer decoder, we sample a small number of visual features and decode visual information under the guidance of language. Finally, a prediction head directly regresses one bounding box for the given referring expression.}
\label{fig:fig_2}
\end{figure*}

\section{Related Work}
\subsection{Visual Grounding}
Visual grounding aims to localize the target object matched with the given referring expression. Existing visual grounding methods can be divided into two-stage methods and one-stage methods.
Early methods address visual grounding task in a two-stage manner, i.e., generating proposals with pre-trained detectors \cite{ren2015faster,he2017mask} and calculating the matching score between each proposal and the referring expression~\cite{mao2016generation,hu2016natural,nagaraja2016modeling,hu2017modeling,zhang2018grounding,yu2018mattnet,liu2019learning,wang2019neighbourhood,yang2019dynamic,liu2020learning}. \cite{nagaraja2016modeling} argues that modeling context between objects is important since referring expressions usually mention relationships of the described object with other objects, and uses multiple instance learning to discover the pairwise context regions. \cite{zhang2018grounding} introduces a variational bayesian method for complex context modeling, avoiding oversimplifying this task to pairwise region modeling. To make use of the structure of the expression, modular networks are proposed to model the compositional linguistic structure of expressions and their described regions by decomposing expressions into modular components \cite{hu2017modeling,yu2018mattnet,liu2019learning}. Some studies further improve two-stage methods with cross modal scene graph for better modeling object relation~\cite{wang2019neighbourhood,yang2019dynamic,liu2020learning}. Although achieving successful performance, two-stage methods heavily depend on the quality of the generated proposals in the first stage and suffer from heavy computation for a large number of proposals. Therefore, there are some works devoted to one-stage visual grounding methods to accelerate the inference speed of models~\cite{yang2019fast,liao2020real,yang2020improving}. One-stage methods usually fuse the visual features and language features first, and then densely regress the bounding box on each position of the feature map grid. Inspired by one-stage object detectors, FAOA~\cite{yang2019fast} fuses the text query’s embedding into the YOLOv3 object detector~\cite{redmon2018yolov3}, and makes a trade-off between accuracy and speed. RCCF~\cite{liao2020real} reformulates visual grounding task as a correlation filtering process by treating the language feature as a dynamic kernel to perform correlation filtering on the image feature map. ReSC~\cite{yang2020improving} proposes a recursive sub-query construction framework for grounding long and complex queries. 

Recently, some studies have proposed transformer based visual grounding methods for fine-grained and sufficient vision-language interaction~\cite{deng2021transvg,kamath2021mdetr,li2021referring,du2021visual,ye2022shifting, yang2022improving,deng2022transvg++}. Specifically, TransVG~\cite{deng2021transvg} adopts a standard multimodal transformer framework consisting of a stack of transformer encoder layers to establish the multimodal correspondence. \cite{kamath2021mdetr,li2021referring,du2021visual} adopt detr-like~\cite{carion2020end} structures to decode bounding boxes from learnable queries. QRNet~\cite{ye2022shifting} proposes query-aware dynamic attention and multi-scale fusion to dynamically compute query-dependent visual attention at the spatial and channel dimensions. TransVG++~\cite{deng2022transvg++} adopts a purely transformer-based framework, removing the multimodal encoder built on top of the backbone, and using a language prompter/adapter in the intermediate layers of the visual backbones for earlier multomodal interaction. VLTVG~\cite{yang2022improving} proposes a visual-linguistic verification module and a language-guided context encoder as multimodal encoder to enhance feature representation, and a query-based multi-stage cross-modal decoder to iteratively update the localization results. The performance improvement of these models mainly comes from designing stronger backbones or multimodal encoders, while our work decouples the visual grounding task into an encoder-decoder framework and improves the performance and efficiency of the model by designing a dynamic decoder, which is orthogonal to the study of these works. Due to the decoupled design of our encoder-decoder, we can easily deploy our decoder on top of these powerful encoders (e.g., QRNet~\cite{ye2022shifting} and TransVG++~\cite{deng2022transvg++}) or use a stronger multimodal pre-trained encoder to further improve model performance.

\subsection{Multimodal Transformers}
Multimodal transformer-based models play an important role in multimodal learning. By pre-training on a large amount of aligned image-text pairs~\cite{sharma2018conceptual} and fine-tuning on datasets of downstream tasks, multimodal transformers show strong performance. These methods can be categorized into single-stream methods~\cite{li2019visualbert,li2020oscar,chen2019uniter,su2019vl} and two-stream methods~\cite{lu2019vilbert,tan2019lxmert,li2021align}. For single-stream methods, the image features and language features are concatenated into a sequence, then the sequence is fed into a multimodal transformer encoder, to model the multimodal interactions via self-attention. Since transformer is permutation-invariant, a learnable positional embedding is added to the input tokens for injecting position information. For two-stream methods, only cross attention is used to model cross-modal interaction. Specifically, one modal serves as query, and the other serves as key and value. Benefiting from self-supervised training strategies (e.g., masked language modeling~\cite{devlin2018bert}, masked region regression~\cite{qi2020imagebert}, and image-text matching~\cite{lin2020interbert}), both kinds of methods achieve great success in many multimodal tasks, such as visual commonsense reasoning~\cite{su2019vl}, visual question answering~\cite{chen2019uniter}, image captioning~\cite{li2020oscar} and visual grounding~\cite{lu2019vilbert}. There are also some works applying multimodal transformer framework into downstream multimodal tasks directly without large-scale multimodal pre-training and still achieveing great performance~\cite{deng2021transvg, du2021visual}, which prove the effectiveness of the multimodal transformer architecture itself.

Although multimodal transformers achieve great success on many downstream tasks, they struggle with dense prediction tasks, such as segmentation and detection, especially when processing high-resolution images, for its huge computational complexity of quadratic of sequence length. This motivates us to design Dynamic MDETR to reduce computational complexity of multimodal transformers.

\subsection{Efficient Transformers}
Self-attention in Transformer~\cite{vaswani2017attention} endows the model with global receptive field and larger modeling capacity, at the cost of huge memory and heavy computation. An effective way towards efficient transformers is to reduce model-level redundancy for efficient computing, such as exploring the low-rank property in self-attention and reducing the self-attention complexity from $O(n^2)$ to $O(n)$~\cite{wang2020linformer,choromanski2020rethinking}.

In this work, we focus on another kind of efficient transformers, i.e., reducing data-level redundancy for efficient computing. In fact, there is spatial redundancy in natural images, such as background areas not consisting of any objects, or objects that have no help for recognizing the target object, which is more severe in images of high resolution. There are some works trying to reduce spatial redundancy for efficient transformers~\cite{rao2021dynamicvit,pan2021ia,song2021dynamic,roh2021sparse}. DynamicViT~\cite{rao2021dynamicvit} progressively prunes redundant tokens based on informative score predicted by MLP. IA-RED$^2$~\cite{pan2021ia} proposes a multi-head interpreter to hierarchically reduce redundant tokens. DGE~\cite{song2021dynamic} proposes a token merging method, using coarse-grained patch splitting for uninformative patches and fine-grained patch splitting for informative patches. Sparse DETR~\cite{roh2021sparse} keeps the image structure, and just selectively updates parts of visual features. There are some differences between these methods and our Dynamic MDETR. First, different from these methods, our proposed Dynamic MDETR samples in the 2D space, i.e., sampling a small number of visual features by predicting x and y offsets relative to the reference point. In addition to different specific techniques, our target tasks are different. Dynamic ViT makes image classification, Sparse DETR is applied to object detection, and DGE validates its effectiveness on image classification, object detection and segmentation. All of these works deal with single modal data (images), while our Dynamic MDETR focuses on multimodal data (images and text). And our work is the first to explore dynamic and efficient framework for visual grounding task. 

We also note that the 2D sampling method of our Dynamic MDETR is similar to Deformable DETR~\cite{zhu2020deformable}. The key difference is that we make careful designs for query generation, to better adapt to visual grounding task. As shown in Figure \ref{fig:fig_4}(b), we do fusion between sampling query (like object query in Deformable DETR) and language features, and sample visual features under the guidance of text. Since injecting strong semantic features, our model can localize target objects more accurately.

\section{Dynamic MDETR}
We formulate the visual grounding task as follows. Given an image $I \in \mathbb{R}^{3 \times H \times W}$ and a referring expression $E=\{w_i\}_{i=1}^L$, where $w_i$ is the $i$-th word in the expression and $L$ is the length of the expression, visual grounding aims to localize the referred object and output the coordinates of the bounding box $b = ({x},{y},{w},{h})$, where $(x,y)$ is the box center coordinates, and $(w,h)$ are width and height of the predicted bounding box.

As shown in Figure \ref{fig:fig_2}, our proposed Dynamic MDETR consists of four main components: a feature encoder, a multimodal transformer encoder, a dynamic multimodal transformer decoder, and a prediction head. We first extract visual and language features from the pre-trained encoders and concatenate them into one sequence. Then we feed the token sequence into the multimodal transformer encoder for cross-modal alignment and fusion, which consists of M vanilla self-attention layers. After the multimodal transformer encoder, we separate the sequence back to visual features and language features. Different from the original DETR~\cite{carion2020end} and MDETR~\cite{kamath2021mdetr}, we directly use the language features as queries instead of learnable object queries. Our dynamic multimodal transformer decoder only picks a small number of sampled feature points via 2D adaptive sampling, and decodes these sampled visual features under the guidance of language queries. Finally, a MLP-based prediction head is used to regress the bounding box.

In this section, we first review the backgrounds in Sec. \ref{sec_3_1}, then detail the techniques of the core components of Dynamic MDETR in Sec.\ref{sec_3_2}-\ref{sec_3_6}.

\subsection{Preliminaries}
\label{sec_3_1}
Before introducing our proposed Dynamic MDETR in details, we provide a brief review on multimodal transformer, DETR~\cite{carion2020end} and MDETR~\cite{kamath2021mdetr} as backgrounds.
\subsubsection{Multimodal Transformer}
Attention mechanism is the core component of transformer~\cite{vaswani2017attention}. Given query embedding $Q \in \mathbb{R}^{M \times C}$, key embedding $K \in \mathbb{R}^{N \times C} $ and value embedding $V \in \mathbb{R}^{N \times C}$, the output of single-head scaled dot-product attention is computed as:
\begin{equation}
    \text{Attention}(Q,K,V) = \text{softmax}(\frac{QK^\mathrm{T}}{\sqrt{d_k}})V \label{eq0}
\end{equation}
For Multi-head attention, the output is the projection of the concatenation of multiple single attention head outputs:
\begin{equation}
\begin{aligned}
    \text{MHA}(Q,K,V) &= \text{Concat}(\text{head}_1, \cdots, \text{head}_H)W^O, \\
     \text{head}_i &= \text{Attention}(QW_i^Q, KW_i^K, VW_i^V),
\end{aligned}
\label{eq:mha}
\end{equation}
where $W$ are learnable projection matrices for feature transformation.

Transformer~\cite{vaswani2017attention} introduces few modal-specific inductive biases and architectural assumptions~\cite{dosovitskiy2020image}, and is compatible with various input modalities, including vision and language. By concatenating visual tokens and textual tokens into a sequence and inputting the sequence into the standard transformer, we can perform self-attention on each token to attend to all other tokens in the sequence. This global receptive field facilitates multimodal information interactions. Recently, many works are devoted to the studies of multimodal transformers. Although multimodal transformers have yielded exciting results on many vision-language tasks, they suffer from heavy computation due to the quadratic computational complexity of input sequence length. To alleviate this problem, we build Dynamic MDETR by replacing computationally expensive encoder layers with our proposed efficient dynamic decoder layers.

\subsubsection{DETR and MDETR Revisited}
DETR~\cite{carion2020end} is an end-to-end query based object detector based on transformer, consisting of a CNN backbone and a transformer encoder-decoder architecture. The core idea of DETR is viewing object detection as a direct set prediction problem and using a fixed number of learnable positional embeddings as object queries in the decoder to decode the objects. Trained with Hungarian matching loss, DETR performs bipartite matching between predicted and ground-truth objects and removes non-maximal suppression. MDETR~\cite{kamath2021mdetr} is an end-to-end modulated multimodal detector based on DETR~\cite{carion2020end}, consisting of a multimodal transformer encoder for multimodal fusion and a query-based transformer decoder to detect all objects. 

We follow the query-based detector, and use language queries to decode the localization of the grounded object. Our proposed Dynamic MDETR makes two important modifications. The first is introducing 2D adaptive sampling to the transformer decoder and enabling the decoder to dynamically sample a small number of visual features, such as 9\% points. With the help of the 2D adaptive sampling, Dynamic MDETR achieves better performance with lower computational cost. The second is that Dynamic MDETR uses language features instead of learnable positional embeddings as queries. Using language as queries, our model can directly sample informative visual features, semantically aggregate visual cues under the guidance of language, and better model relationships between images and language descriptions.

\subsection{Feature Encoder}
\label{sec_3_2}
Following TransVG~\cite{deng2021transvg}, we use ResNet~\cite{he2016deep} followed by a transformer encoder or ViT~\cite{dosovitskiy2020image} as visual backbone. The visual encoder is initialized with the backbone and transformer encoder of DETR~\cite{carion2020end}. For ViT-B backbone, we use the CLIP pre-training. For language encoder, we use the uncased version of Bert~\cite{devlin2018bert} or the CLIP pre-trained text encoder. For the given image-expression pair $<I, E>$, the output of the visual encoder is $Z_v \in \mathbb{R}^{N_v \times C_v}$ and the output of the language encoder is $Z_l \in \mathbb{R}^{N_l \times C_l}$, where $N_{\{v,l\}}$ represents the token number and $C_{\{v,l\}}$ represents the channel number. One FC layer for each modal is used to project them into the same embedding space with the same number of channels, thus we can get visual features $F_v \in \mathbb{R}^{N_v \times C}$ and language features $F_l \in \mathbb{R}^{N_l \times C}$. Then we concatenate them into a sequence $F = [F_v;F_l] \in \mathbb{R}^{(N_v + N_l) \times C}$, as the input of the multimodal transformer encoder.

\begin{figure}
\centering
\includegraphics[width=1.0\linewidth]{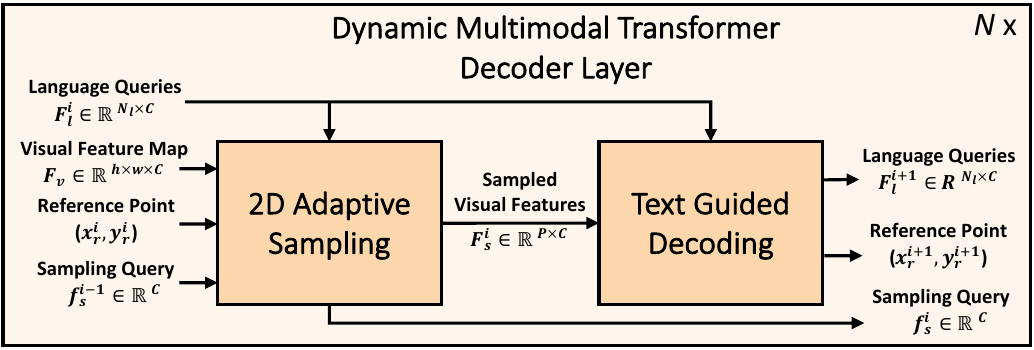}
\caption{Dynamic multimodal transformer decoder is a stack of N layers. Each decoder layer consists of two submodules: 2D adaptive sampling and text guided decoding. The former samples a small number of visual points in the 2D image space by predicting offsets relative to the reference point, and the latter adopts one-layer transformer encoder-decoder architecture to decode the sampled visual features under the guidance of language for subsequent processing.}
\label{fig:fig_3}
\end{figure}

\begin{figure*}
\centering
\includegraphics[width=\linewidth]{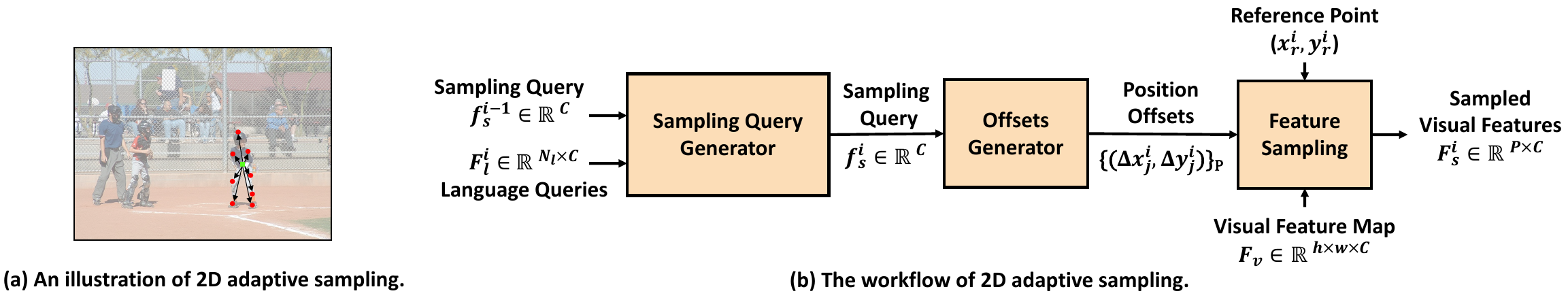}
\caption{The process of {\bf 2D Adaptive Sampling}. (a) An illustration of 2D adaptive sampling. Our sampling method samples visual features in the 2D space by predicting offsets on x axis and y axis relative to the reference point, to better capture the positional and geometric information of the target object. (b) The workflow of 2D adaptive sampling. 2D adaptive sampling consists of three steps, including sampling query generator, offsets generator and feature sampling module. Specifically, we first fuse the sampling query from the last decoder layer and language queries in the sampling query generator to get the language guided sampling query. Then we do a simple linear projection on the sampling query to generate the offsets relative to the reference point. In the feature sampling module, we calculate the absolute coordinates of the sampled points. Since the coordinates are real values, we do bilinear interpolation on the feature map and get the sampled visual features.}
\label{fig:fig_4}
\end{figure*}

\subsection{Multimodal Transformer Encoder}
\label{sec_3_3}
Multimodal Transformer Encoder has M standard transformer encoder layers~\cite{vaswani2017attention}, consisting of a multi-head self-attention module and a feed forward network. The input of multimodal transformer encoder is the concatenated visual and language feature sequence $F$. Since the transformer architecture is permutation-invariant, we add a learnable positional embedding $P \in \mathbb{R}^{(N_v+N_l)\times C}$ to the input of each encoder layer like DETR~\cite{carion2020end}. The calculation of the $i$-th layer can be formulated as:
\[
    \begin{aligned}
    Q_E^i &= K_E^i = V_E^i  = F^{i-1},\\
    \hat{F}^i &= \text{LN}(Q_E^i + \text{MHA}(Q_E^i + P, K_E^i + P, V_E^i)), \\
    F^i &= \text{LN}(\hat{F^i} + \text{FFN}(\hat{F}^i)),
\end{aligned}
\]
where MHA$(\cdot)$  means multi-head attention in Equation (\ref{eq:mha}), FFN$(\cdot)$ is the feed forward network and LN$(\cdot)$ denotes layer normalization~\cite{ba2016layer}. $F^{i-1}$ is the output sequence of the $i-1$-th encoder layer and the input sequence of the $i$-th encoder layer, and $F^0 = F$. The output of the $i$-th encoder layer is $F^i$ and the output of the multimodal transformer encoder is denoted as $F_E$, where $F_E = F^M$. Via self-attention, multimodal transformer encoder can model intra-modal and inter-modal interactions effectively.

After the multimodal transformer encoder, we split the output $F_E$ to visual features $F_v \in \mathbb{R}^{N_v \times C}$ and language features $F_l \in \mathbb{R}^{N_l \times C}$, and split the positional embedding $P$ to visual positional embedding $P_v \in \mathbb{R}^{N_v \times C}$ and language positional embedding $P_l \in \mathbb{R}^{N_l \times C}$. Then we input the features and positional embeddings to our dynamic multimodal transformer decoder for efficient localization of the grounded object.

\subsection{Dynamic Multimodal Transformer Decoder}
\label{sec_3_4}
Although powerful, multimodal transformer encoders suffer from heavy computation due to the standard self-attention operation with quadratic time complexity. Therefore, we introduce the dynamic sampling to the design of multimodal transformers and propose a dynamic multimodal transformer decoder. With the help of the dynamic sampling, our decoder only needs a small number of discriminative visual features, which eliminates the decoding complexity problem of depending on the image input size. As shown in Figure \ref{fig:fig_3}, our dynamic multimodal transformer decoder has $N$ layers. Each decoder layer consists of two submodules: 2D adaptive sampling and text guided decoding. The former samples a small number of spatial points on the 2D feature map and the latter decodes these sampled visual features for the grounding task.

\subsubsection{2D adaptive sampling}
As shown in Fig.4 (a), our proposed 2D adaptive sampling module samples visual features in the 2-dimensional image space, via predicting offsets (black lines with arrow) relative to the given reference point (green point). As the decoder layer goes deeper, the sampled points will gradually concentrate on the target object, falling inside or at the boundary. Therefore, our 2D adaptive sampling method is able to extract discriminative points around grounded objects and thus better capture geometric information for accurate regression.

Specifically, our 2D adaptive sampling consists of three steps: a {\em sampling query generator}, a {\em offsets generator}, and a {\em feature sampling module}. Taking the $i$-th decoder layer as an example, for the sampling query generator, we first perform average pooling on language queries and get the language query $f_l^i \in \mathbb{R}^{C}$. We then concatenate the sampling query $f_s^{i-1} \in \mathbb{R}^{C}$ from the last decoder layer and the language query $f_l^i \in \mathbb{R}^C$, and feed it into a two-layer MLP to generate a new language-guided sampling query $f_s^i \in \mathbb{R}^{C}$ for the subsequent offsets generation. The calculation process is as follows:
\[
    \begin{aligned}
        &f_s^i = \text{MLP}([f_s^{i-1}; f_l^i]).\\
    \end{aligned}
\]
The concatenation and fusion of two queries are to preserve the foreground information and inject the guidance of text for better feature sampling and localization. Then we input the sampling query $f_s^i$ to the offset generator. More concretely, we use a simple linear projection layer to generate the offsets relative to the reference point: 
\begin{align}
    \{(\Delta x_j^i, \Delta y_j^i)\}_{j=1}^P = {\rm Linear}(f_s^i),
\end{align}
where $P$ is the number of sampled points, and $(\Delta x, \Delta y)$ are the predicted offsets relative to the reference point. In the feature sampling module, we first generate the absolute sampled positions. Given the reference point $(x_r^i, y_r^i)$, we can get the sampled locations:
{\small
\begin{align}
    \begin{cases}
        x_j^i = x_r^i + \Delta x_j^i, \\
        y_j^i = y_r^i + \Delta y_j^i.
    \end{cases}
\end{align}
}
With the coordinates of the sampled points, we perform bilinear interpolation on the visual feature map to get the sampled visual features for the subsequent text guided decoding. The sampled visual features are denoted as $F_s^i \in \mathbb{R}^{P \times C}$. Similarly, we can also sample the corresponding positional embeddings $P_s^i \in \mathbb{R}^{P \times C}$.

For initialization, the input reference point is initialized as the normalized coordinate $(0.5, 0.5)$, i.e., the center of the image, the sampling query is a randomly initialized learnable vector, and the language queries are from the multimodal transformer encoder. For the first dynamic decoder layer, the sampling query generator only takes the sampling query as input without fusion with the language queries, for input-independent initial sampling of visual features. For the subsequent dynamic decoder layers, these queries are updated iteratively as the outputs of the last decoder layer.

\begin{figure}[t]
\centering
\includegraphics[width=0.9\linewidth]{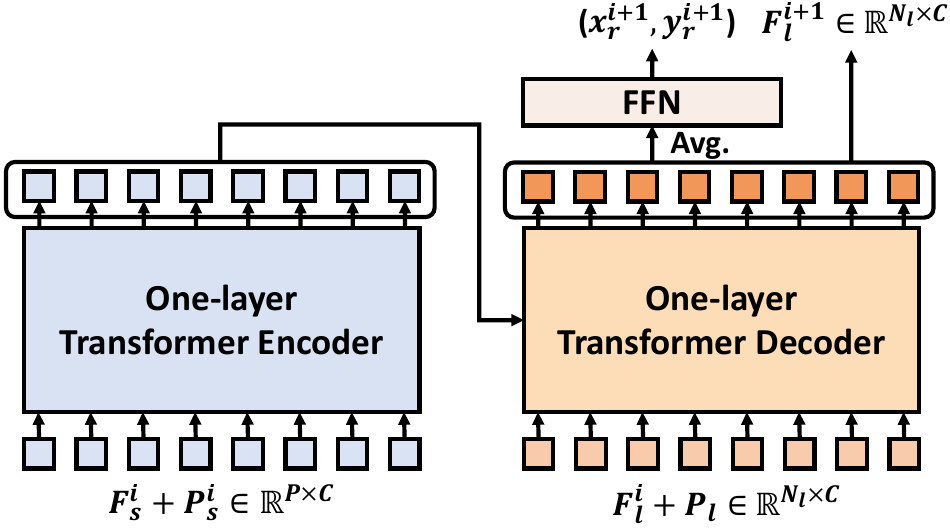}
\caption{The process of {\bf Text Guided Decoding}. After sampling a small number of informative visual features, we use language as queries to decode the visual features under the guidance of language with a one-layer transformer encoder-decoder. In the decoding process, the language queries and reference point are iteratively updated for better feature sampling and visual grounding.}
\label{fig:fig_5}
\end{figure}

\subsubsection{Text guided decoding}
After 2D adaptive sampling, these sampled features $F_s^i$ with their positional embedding $P_s^i$ will be fed into the text guided decoding module to accomplish the task of visual grounding. Our text guided decoding module will directly regress the bounding box coordinates of the grounded object. 

As shown in Figure~\ref{fig:fig_5}, for the $i$-th dynamic multimodal decoder layer, the input of the encoder is the sampled visual features $F_s^i \in \mathbb{R}^{P \times C}$ and the positional embedding $P_s^i \in \mathbb{R}^{P \times C}$. The encoder aims to extract context information among sampled visual features with self-attention operations to enhance their representation power. The decoder inputs are the contextualized representation and the language queries $F_l^i \in \mathbb{R}^{N_l \times C} $ with their positional embedding $P_l$. The decoder employs a cross-attention operation to decode the location of grounded object by using language as queries. The cross-attention operation can gradually align between two modalities, and progressively distill the corresponding visual information for accurate regression. Specifically, the whole pipeline of text guided decoding could be formulated as follows. For the encoder, the computation is formulated as:
\[
\begin{aligned}
    &Q_{DE}^i = K_{DE}^i = V_{DE}^i = F_s^i,\\
    &\hat{F}_s^i = \text{LN}(Q_{DE}^i + \text{MHA}(Q_{DE}^i + P_s^i, K_{DE}^i + P_s^i, V_{DE}^i)), \\
    &\hat{F}_s^i= \text{LN}(\hat{F}_s^i + \text{FFN}(\hat{F}_s^i)),
\end{aligned}
\]
where MHA$(\cdot)$  is the multi-head attention in Equation (\ref{eq:mha}), FFN$(\cdot)$ is the feed forward network and LN$(\cdot)$ denotes layer normalization. For the decoder, the computation is formulated as:
\[
\begin{aligned}
    &K_{DD}^i = V_{DD}^i = \hat{F_s^i}, Q_{DD}^i = F_l^i, \\
    &\hat{F}_l^{i+1} = \text{LN}(Q_{DD}^i + \text{MHA}(Q_{DD}^i + P_l, K_{DD}^i + P_s^i, V_{DD}^i)), \\
    &F_l^{i+1} = \text{LN}(\hat{F}_l^{i+1} + \text{FFN}(\hat{F}_l^{i+1})).
\end{aligned}
\]
After decoder layer, the queried representation is able to encode the aligned information between text and the corresponding grounded object.  We will use this representation to update the reference point after each decoder layer, which will gradually converge to the object center. More concretely, the updated language queries are averaged pooling to one vector, and then we feed it into a feed forward network to predict the new reference point. Meanwhile, the language query will be updated with the queried representation to progressively bridge the modality gap between image and text. Overall, as shown in Figure \ref{fig:fig_3}, our dynamic multimodal decoder is built by stacking multiple layers of 2D adaptive sampling and text guided decoding. This progressive decoding process will gradually bridge the modality gap and iteratively refine the bounding box center, eventually realizing the objective of visual grounding.

\begin{figure}[t]
\centering
\includegraphics[width=\linewidth]{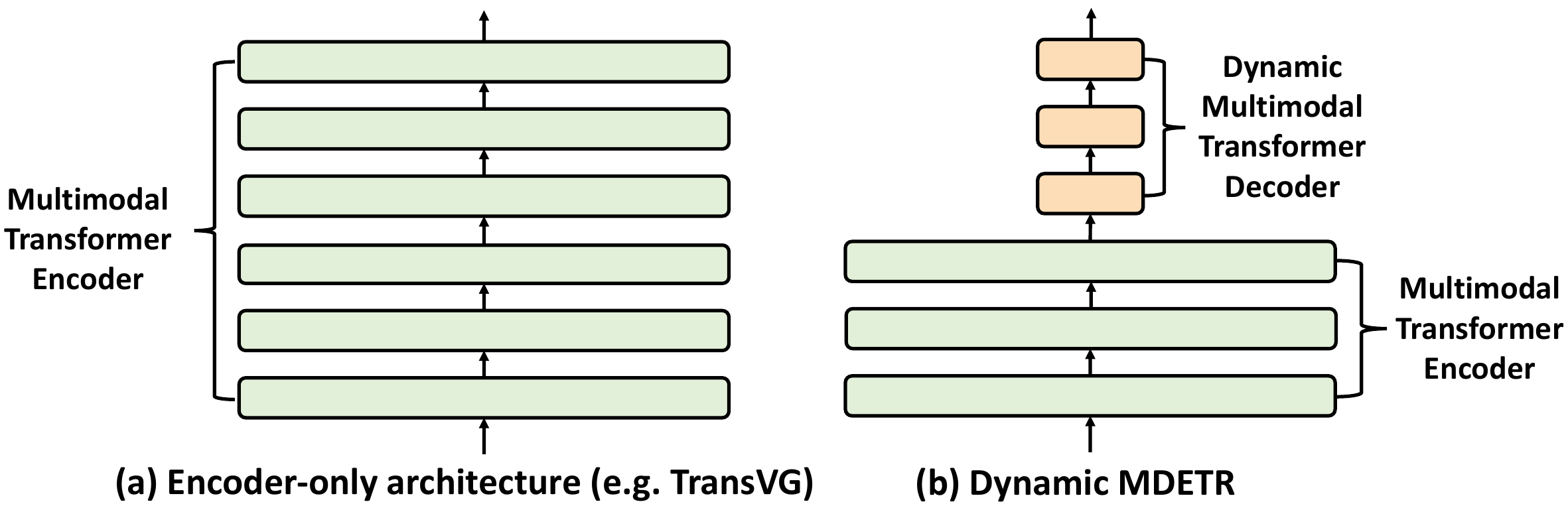}
\caption{Illustration of TransVG and Dynamic MDETR. (a) TransVG has a six-layer multimodal transformer encoder. For each encoder layer, all feature points in the feature map are used. (b) Dynamic MDETR has an M-layer multimodal transformer encoder and an N-layer dynamic multimodal transformer decoder. For the default setting, we set both M and N to 3. Compared with vanilla multimodal transformer encoder layer, our proposed dynamic multimodal transformer layer only uses a small number of sampled visual feature points, such as 9\% points, thus can reduce a large number of computations.}
\label{fig:fig_6}
\end{figure}
\subsubsection{Discussion on modeling efficiency}
Multimodal transformer exhibits high capacity and flexibility to align image and text for visual grounding task. As shown in the left of Figure~\ref{fig:fig_6}, a simple baseline method (e.g. TransVG~\cite{deng2021transvg}) is to directly use the multimodal transformer encoder to operate on the concatenated token sequence of image and text. This pure encoder architecture could be effective on capturing the correlation between two modalities and bridging their modality gap. However, this architecture is quite computationally expensive as it involves both intra-modality self attention and inter-modality cross attention. Instead, our dynamic MDETR proposes to decouple the visual grounding process into encoding and decoding phases. In the encoding phase, we mainly focus on the alignment of feature space between two modalities while in the decoding phase, we pay attention to the exact visual localization of the grounded object described by the text. More importantly, in this decoupled architecture, we can devise customized and more efficient decoding strategy to handle the sparsity problem of visual grounding (i.e., only a small number of visual tokens contribute to the final prediction). Via 2D adaptive sampling, our dynamic multimodal transformer decoder uses a small number of sampled points, such as 36 points as input, which is far less than 400 points used in the multimodal transformer encoder. As shown in the right of Figure~\ref{fig:fig_6}, the dynamic multimodal transformer decoder layer with fewer points requires lower computational cost. By replacing some multimodal transformer encoder layers with dynamic multimodal transformer decoder layers, our Dynamic MDETR can reduce a large number of computations. As shown in experiments, our Dynamic MDETR achieves comparable or better results, but far fewer flops than TransVG~\cite{deng2021transvg}, which adopts an encoder-only multimodal transformer architecture.

\subsection{Prediction Head}
\label{sec_3_5}
Given the output of the dynamic multimodal transformer decoder $F_D \in \mathbb{R}^{N_l \times C}$ and padding mask $m_l \in \mathbb{R}^{N_l}$, we calculate the average of the unmasked tokens as $f_{reg} \in \mathbb{R}^C$. Then we fed $f_{reg}$ into the prediction head consisting of 3 fully-connected layers with ReLU activation function. The prediction head directly regresses 4-dim bounding box coordinates:
\[
\hat{b} = (\hat{x},\hat{y},\hat{w},\hat{h}) = \text{MLP}(f_{reg}),
\]
where $(\hat{x},\hat{y})$ are the normalized center coordinates, and $(\hat{w},\hat{h})$ are the width and height of the predicted bounding box. 
\subsection{Loss Function}
\label{sec_3_6}
We cast the visual grounding task as a direct regression task, which means we only predict one bounding box for each referring expression without needing to select the final result from many candidates. As a result, we can remove the classification branch.

To optimize our model, we adopt the combination of the L1 loss and the scale-invariant generalized IoU loss~\cite{rezatofighi2019generalized} as our optimization objectives. Specifically, we denote the predicted bounding box as $\hat{b} = (\hat{x},\hat{y},\hat{w},\hat{h})$ and the target box as $b = (x,y,w,h)$. The loss function is defined as:
\[
\mathcal{L} = \mathcal{L}_{L1}(b, \hat{b}) +  \mathcal{L}_{GIoU}(b, \hat{b}),
\]
where $\mathcal{L}_{L1}(\cdot)$ and $\mathcal{L}_{GIoU}(\cdot)$ are the L1 loss and GIoU loss, respectively. 

\begin{figure*}[t]
    \centering
    \includegraphics[width=1.0\linewidth]{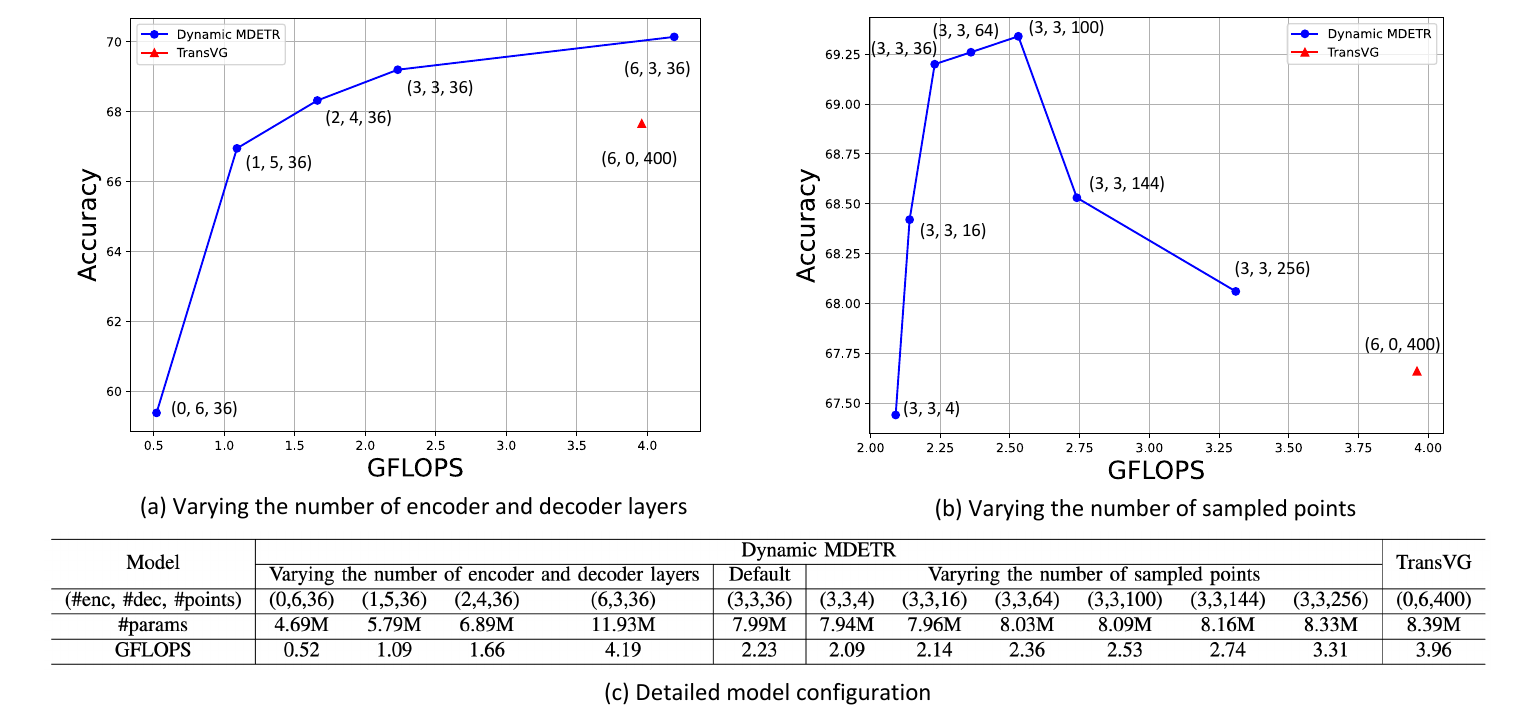}
    \caption{FLOPs and accuracy trade-offs of multimodal transformer models on validation set of RefCOCOg-umd. (a) Keep the number of sampled points and vary the numbers of the encoder and decoder layers. (b) Keep the numbers of the encoder and decoder layers and vary the number of sampled points. We compare our Dynamic MDETR with state-of-the-art multimodal transformer model TransVG. (c) Detailed model configurations. (\#enc, \#dec, \#points) denotes the number of multimodal transformer encoder layers, dynamic multimodal decoder layers and sampled points that the model has. The default setting for Dynamic MDETR is (3,3,36), i.e., 3 encoder layers, 3 decoder layers and 36 sampled points. TransVG can be regarded as having six encoder layers and zero decoder layers, and using all the points of the feature map, i.e., 400 points.}
    \label{fig:fig_7}
\end{figure*}

\section{Experiments}
In this section, we conduct extensive experiments to verify our Dynamic MDETR on five common benchmarks of visual grounding. Specifically, we first introduce the datasets and standard evaluation protocol in Sec. \ref{sec_4_1}. Then we detail the model implementation, including hyper-parameters and training recipes in Sec. \ref{sec_4_2}. After this, we first perform some ablation studies to investigate the effect of each component of Dynamic MDETR in Sec. \ref{sec_4_3}. In Sec. \ref{sec_4_4} we show the main results and compare with the state-of-the-art methods to demonstrate the effectiveness of our method. We also transfer our method to VQA task to prove its generalization to other vision-language domain in Sec. \ref{sec_4_5} Finally, we visualize some examples for an intuitive explanation of the learning process in Sec. \ref{sec_4_6}.

\subsection{Datasets and Evaluation Metric}
\label{sec_4_1}

We conduct extensive experiments on five datasets. The details about these datasets are as follows:

\textbf{RefCOCO/RefCOCO+/RefCOCOg.} RefCOCO~\cite{yu2016modeling}, RefCOCO+~\cite{yu2016modeling} and RefCOCOg~\cite{mao2016generation} are collected from MSCOCO~\cite{lin2014microsoft}. Specifically, RefCOCO includes 19,994 images with 142,210 referring expressions for 50,000 referred objects, RefCOCO+ includes 19,992 images with 141,564 referring expressions for 49,856 referred objects, and RefCOCOg includes 25,799 images with 95,010 referring expressions for 49822 referred objects. Following the split used in \cite{deng2021transvg}, we report the performance on the validation, testA (containing multiple people) and testB (containing multiple instances of other objects) splits for RefCOCO and RefCOCO+, validation split for RefCOCOg-google~\cite{mao2016generation}, validation and test splits for RefCOCOg-umd~\cite{nagaraja2016modeling}.

\textbf{ReferItGame.} ReferItGame~\cite{kazemzadeh2014referitgame} includes 20,000 images with 120,072 referring expressions for 19,987 referred objects. We follow \cite{deng2021transvg} to split the dataset into train, validation and test set, and report the performance on the test set.

\textbf{Flickr30K Entities.} Flickr30K Entities~\cite{plummer2015flickr30k} includes 31,783 images with 427,000 referring expressions for 427,000 referred objects. We follow \cite{deng2021transvg} to split the images into 29,783 for training, 1000 for validation, and 1000 for testing, and report the performance on the test set.

\textbf{Evaluation Metric.}
We follow the standard metric to report top-1 accuracy (\%), where the prediction is correct if the IoU between the predicted box and ground-truth box is above 0.5.

\subsection{Implementation Details}
\label{sec_4_2}
Following TransVG~\cite{deng2021transvg}, we keep the aspect ratio of the input image, resize the longer edge to 640 and pad the shorter edge to 640 with the mean value of RGB channels. Example images are shown in Figure~\ref{fig:fig_1}. The maximum expression length is set to 40 for RefCOCOg and 20 for other datasets. We cut off the referring expression if its length exceeds the max length, otherwise we pad it to the max length. By default, our visual encoder is initialized with the ResNet backbone~\cite{he2016deep} and transformer encoder~\cite{vaswani2017attention} of DETR model~\cite{carion2020end}, and our language encoder is initialized with BERT of the uncased version~\cite{devlin2018bert}, which are the same as TransVG. 
We set the number of multimodal encoder layers and dynamic multimodal decoder layers to 3, and the number of sampled points to 36 as the default setting. 

For training, we use AdamW~\cite{loshchilov2017decoupled} to optimize our model for 90 epochs, and set weight decay to 10$^{-4}$ and dropout ratio to 0.1. The initial learning rate is set to 10$^{-5}$ for pre-trained feature encoders and 10$^{-4}$ for other parameters, with a drop by a factor of 10 at 60 epochs on RefCOCO/RefCOCO+/RefCOCOg and ReferitGame. On Flickr30K Entities, we train our model for 60 epochs, with a learning rate drop by a factor of 10 at 40 epochs. For Referit Game, we find that despite a smaller dataset size, our model shows no sign of overfitting. Therefore, we train our model on ReferitGame without weight decay and dropout. The batch size is set to 128. We also follow TransVG to perform the data augmentation for training.

In addition, to further demonstrate the generalization ability of our Dynamic MDETR framework, we also use the ViT-B/16 CLIP encoder~\cite{radford2021learning} whose visual model and language model are jointly trained on matched image-text pairs. We set the number of multimodal encoder layers and dynamic multimodal decoder layers to 3, and the number of sampled points to 160 (1600 in total) as the default setting. We train our model for 60 epochs with a learning rate drop by a factor of 10 at epoch 45. The initial learning rate is set to 5$\times$10$^{-6}$ for pre-trained feature encoders and 10$^{-4}$ for other parameters. To make the training more stable, we freeze the CLIP encoder in the first 10 epochs. We also remove dropout and weight decay on ReferItGame~\cite{kazemzadeh2014referitgame} and train the model for 90 epochs with a learning rate drop after 60 epochs, but don't freeze the CLIP. For Flickr30K Entities, we train the model for 40 epochs, with a learning rate drops after 30 epochs, and freeze CLIP in the first 6 epochs. The batch size is set to 64. It is worth noting that our Dynamic MDETR with CLIP as backbone does not use any necks like the DETR encoder used in TransVG and is not pre-trained on COCO~\cite{lin2014microsoft}.

\begin{figure*}[t]
\centering
\includegraphics[width=\linewidth]{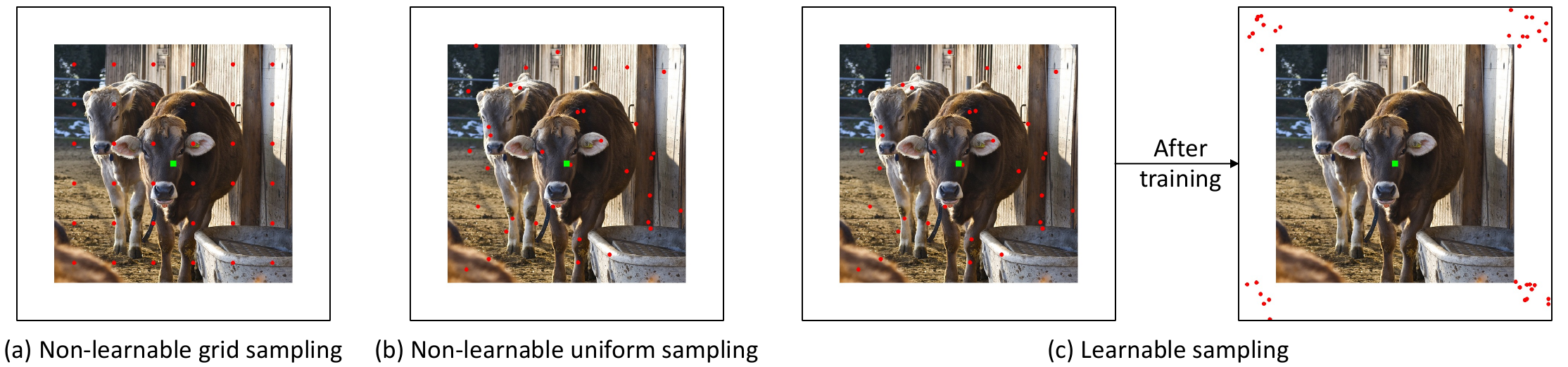}
\caption{Illustrations of initial sampling strategies. For the initial sampling of the first decoder layer, we implement three strategies, including non-learnable grid sampling, non-learnable uniform sampling and learnable sampling initialized by uniform sampling, which are shown in Figure Re 1. Specifically, we sample points on the feature map at equal intervals for both x-axis and y-axis for non-learnable grid sampling. For the other two methods, we use a linear layer, whose weight is zeroed and bias is uniformly drawn in [-0.5, 0.5], to generate offsets relative to the initial reference point (0.5, 0.5) based on the sampling query. The linear layer is fixed for non-learnable uniform sampling, while trainable for learnable sampling.}
\label{fig:initial_sampling}
\end{figure*}

\subsection{Ablation Studies}
\label{sec_4_3}
In this section, we conduct ablation studies on the design of Dynamic MDETR (ResNet-50) on the validation set of RefCOCOg-umd to demonstrate the effectiveness of our proposed dynamic multimodal transformer decoder. 

\textbf{Accuracy and computation trade-offs.} 
To verify the efficiency of our proposed Dynamic MDETR, we perform experiments on trade-offs between accuracy and computation. It is worth noting that the computations of backbones are not included. Specifically, we vary the number of encoder and decoder layers and the number of sampled points to control the model complexity, and compare with a simple but powerful transformer encoder-only baseline method TransVG~\cite{deng2021transvg}. The results are shown in Fig. \ref{fig:fig_7}.

First, we perform experiments on the trade-off between encoder and decoder in our Dynamic MDETR. We fix the total number of encoder and decoder layers at 6 and vary their specific layer configurations. Fig. \ref{fig:fig_7} (a) shows that the more encoder layers can contribute to higher accuracies. Yet, the computational cost also increases with more encoder layers. We can replace the expensive multimodal transformer encoder layers with our proposed efficient dynamic multimodal transformer decoder layers for favorable trade-offs between computation and accuracy. Keeping only 1 encoder layer, Dynamic MDETR achieves 66.95\%,  which is 0.71\% lower than the baseline, but only has 27\% of its GFLOPs. In fact, we find that using the architecture of 3 encoders and 3 decoders can achieve 1.54\% absolute accuracy increase and 44\% relative flops drop compared with the baseline. This choice obtains a good trade-off between accuracy and efficiency, and is used as our default setting in the remaining experiments. We also perform a study by directly adding a three-layer decoder on top of the baseline method. It achieves a better accuracy of 70.14\%, but only brings marginal extra computational cost.

Then, we perform experiments on the effect of the number of sampled points. Fig. \ref{fig:fig_7} (b) illustrates the results by varying the number of sampled points in the 2D sampling module. Surprisingly, Dynamic MDETR achieves a similar accuracy to the baseline by only using 1\% points (i.e., 4 points with 0.22\% top-1 accuracy drop). With more sampled points, Dynamic MDETR outperforms TransVG by a large margin, but still with fewer flops. We also find that not the more sampled points, the better the model performs. Dynamic MDETR achieves the best accuracy with 100 sampled points and the accuracy decreases when sampling more points, which demonstrates there is severe spatial redundancy in images. The severe spatial redundancy not only brings huge and unnecessary computations, but is also harmful to the model performance. By introducing dynamic multimodal decoder, our method can reduce spatial redundancy effectively and efficiently.

\textbf{Ablation on key designs of the dynamic multimodal decoder.}
We begin the ablative experiments on the two key components of the dynamic multimodel decoder to verify its effectiveness: (1) decoupling the visual grounding head into the separate design of encoder and decoder (2) introducing the dynamic design in the decoder. The results are shown in Table \ref{table:table_1}. The baseline model is an encoder-only architecture of TransVG~\cite{deng2021transvg} with a six-layer multimodal transformer encoder and without any decoder. The other models are the ablative versions of our proposed Dynamic MDETR, which has a three-layer multimodal transformer encoder and a three-layer multimodal transformer decoder. Our first variant is to use a three-layer static multimodal decoder with all visual features as input, which removes the 2D adaptive sampling module from the standard Dynamic MDETR. The accuracy of this variant is 69.12\%, which is better than the baseline (67.66\%) with a very similar computation cost (4.10 GFLOPS vs. 3.96 GFLOPS). This demonstrates that decoupling the grounding process into encoder and decoder design is more effective for the downstream localization task, since the decoder can explicitly attend to the useful visual information from the whole feature map and aggregate the sampled visual features under the guidance of the language for better localization. 

Simply using 2D adaptive sampling and replacing the proposed text guided decoding module with a simple average pooling, we achieve slightly higher performance than baseline, by only using 9\% visual feature points and with only around 50\% GFLOPS. This simple method demonstrates that there is severe spatial redundancy in the image domain. Finally, our complete dynamic multimodal decoder achieves 69.20\%, leading to a 1.54\% accuracy rise and 44\% GFLOPS drop compared with the encoder-only baseline. This demonstrates the overall effectiveness and efficiency of our Dynamic MDETR by progressively sampling a small number of informative feature points and performing text guided decoding on these sampled feature points.

\begin{table*}[h]
\centering
\caption{Ablation on the 2D adaptive sampling and text guided decoding in our dynamic multimodal transformer decoder.}
\scalebox{1.0}{
\begin{tabular}{c|c|c|c|c|c}
\hline
Architecture & 2D Adaptive Sampling      & Text Guided Decoding    & \#points     & GFLOPS & Accuracy \\ \hline
Encoder only (baseline) &                          &                       & 400      &   3.96     &    67.66      \\
Encoder-Decoder &                         & \checkmark & 400 & 4.10   & 69.12    \\
Encoder-Decoder&  \checkmark &                           & 36 & 1.99   & 67.89         \\
Encoder-Decoder &\checkmark & \checkmark & 36 & 2.23   & \textbf{69.20}    \\ \hline
\end{tabular}
\label{table:table_1}
}
\end{table*}

\begin{table*}[t]
\parbox{.33\linewidth}{
\centering
\caption{Ablation on initial sampling strategies.}
\scalebox{1.0}{
\begin{tabular}{c|c}
\hline
Method     & Accuracy \\ \hline
Grid      & 67.89    \\
Uniform   & 67.77    \\
Learnable & \textbf{69.20}    \\ \hline
\end{tabular}
\label{table:table_2}
}
}
\hfill
\parbox{.33\linewidth}{
\centering
\caption{Dynamic sampling vs. Static sampling.}
\begin{tabular}{c|c}
\hline
Method     & Accuracy \\ \hline
Static-Uniform     & 66.81    \\ 
Static-Max pooling & 67.85    \\ 
Dynamic-Ours    & \textbf{69.20}    \\ \hline
\end{tabular}
\label{table:table_3}
}
\hfill
\parbox{.33\linewidth}{
\centering
\caption{Ablation on query design.}
\begin{tabular}{c|c}
\hline
Sampling Query      & Accuracy \\ \hline
w/o Language Guided~\cite{zhu2020deformable}     & 68.32    \\ 
w Language Guided & \textbf{69.20}    \\ \hline
\end{tabular}
\label{table:table_4}
}
\end{table*}

\begin{table}[t]
\centering
\caption{Ablation on gradually reducing sampled points.}
\begin{tabular}{c|c|c}
\hline
\#points    & GFLOPS & Accuracy \\ \hline
16-16-16    & 2.14   & 68.42    \\
36-36-36    & 2.23   & 69.20    \\ 
100-100-100 & 2.53   & 69.34    \\ \hline
36-16-9     & 2.16   & 67.99    \\
64-32-16    & 2.24   & 69.34    \\
128-64-36   & 2.38   & 69.38    \\
200-100-50  & 2.62   & 69.42    \\ \hline
\end{tabular}
\label{table_5}
\end{table}

\begin{table}[t]
\centering
\caption{Ablation on decoding strategies.}
\begin{tabular}{c|c|c|c}
\hline
Decoding  & Sampling & GFLOPS & Accuracy \\ \hline
Object Queries (MDETR~\cite{kamath2021mdetr})  & $\times$ & 2.66 & 66.81    \\ \hline
Average Pooling & $\checkmark$ & 1.99 & 67.89    \\ 
Self Attention & $\checkmark$ & 2.21 & 68.30    \\ 
Text Guided Decoding  & $\times$ & 4.10 & 69.12    \\
Text Guided Decoding  & $\checkmark$& 2.23 & \textbf{69.20}    \\ \hline
\end{tabular}
\label{table:table_6}
\end{table}

\begin{table}[t]
\centering
\caption{Comparison with other dynamic visual grounding transformers. 'P' means 'Progressive' and 'TGD' denotes 'Text Guided Decoding'. }
\resizebox{\linewidth}{!}{
\begin{tabular}{c|c|c|c|c}
\hline
Method    & Sampling   & GFLOPS  & Latency & Accuracy \\ \hline
\textbf{\textit{Encoder only:}} &   &   &    & \\
TransVG  & No & 3.96    & 19 ms  & 67.66    \\
TransVG-Random & 1D   & 2.22  & 12 ms   & 66.12    \\ 
TransVG-Random-P & 1D  & 2.75   & 16 ms   & 66.42    \\
TransVG-Attention & 1D  & 2.22  & 12 ms  & 67.67    \\ 
TransVG-Attention-P & 1D  & 2.75 & 16 ms & 68.34    \\ \hline % 2.75
\textbf{\textit{Encoder-Decoder:}} &       & \\
TransVG + our TGD & No     & 4.10   & 21 ms   & 69.12 \\
Dynamic MDETR (Ours)  & 2D     & 2.23 & 14 ms    & 69.20    \\
Dynamic MDETR-P (Ours)  & 2D     & 2.38 & 15 ms    & \textbf{69.38}   \\ \hline % 2.23
\end{tabular}
}
\label{table:table_7}
\end{table}

\textbf{Ablation on initial sampling strategies.}
We also investigate the impact of different initial sampling strategies before inputting to the decoder, including non-learnable grid sampling, non-learnable uniform sampling and learnable sampling initialized by uniform sampling. As shown in Fig. \ref{fig:initial_sampling}, we sample points on the feature map at equal intervals for both x-axis and y-axis for non-learnable grid sampling. For the other two methods, we use a linear layer, whose weight is zeroed and bias is uniformly drawn in $[-0.5, 0.5]$, to generate offsets relative to the initial reference point $(0.5, 0.5)$ based on the sampling query. In fact, this sampling process is 2D adaptive sampling in the first dynamic decoder layer. The linear layer is fixed for non-learnable uniform sampling, while trainable for learnable sampling. Table \ref{table:table_2} shows that the learnable sampling is the best. We infer that learnable initial sampling can learn a better location distribution of objects, thus performing better. We use it as our default initial sampling strategy.

\textbf{Comparisons between dynamic sampling and static sampling.}
Our proposed 2D adaptive sampling method is conditioned on the input image and text description. It adaptively samples visual features from the 2D feature map based on the visual content and language query. To discuss whether this dynamic sampling is necessary, we design two 2D static sampling methods, including uniform sampling and max pooling based sampling. Specifically, we sample 36 visual tokens from a uniform distribution for the first static sampling method, and downsample the original feature map to 6$\times$6 by max pooling for the second static sampling method. The results are shown in Table \ref{table:table_3}. We find that our proposed 2D adaptive sampling outperforms the 2D static sampling methods by a large margin, thanks to the modeling flexibility of our Dynamic MDETR.

\textbf{Ablation on Sampling Query Design.} As illustrated in Figure \ref{fig:fig_4}(b), we do fusion between sampling query (like object query in Deformable DETR~\cite{zhu2020deformable}) and language features. When removing the fusion and only using the sampling query, our sampling method degenerates to the method of Deformable DETR~\cite{zhu2020deformable}. To discuss whether generating language guided sampling query is better, we conduct ablation experiments and report the results in Table \ref{table:table_4}. The results show that ours is better, and we infer that sampling visual features under the guidance of text can inject strong semantic features and leads to more accurate localization. 

\textbf{Ablation on gradually reducing sampled points.} To further improve dynamic capability, we progressively decrease the number of sampling points in the decoder and show the results in Table \ref{table_5}. Compared "64-32-16" with "36-36-36", "64-32-16" and "128-64-32" with "100-100-100", we can find that this sampling strategy of gradually reducing the number of sampling points outperforms sampling the same number of points in each decoder layer, i.e., achieving higher accuracy with less or similar computations.

\textbf{Ablation on decoding strategies.} We compare our text guided decoding with several common decoding strategies and report the results in Table \ref{table:table_6}. The decoding strategies include: performing average pooling on sampled visual features (\textit{Average Pooling}), performing self-attention operation on the sequence of sampled visual features and text features (\textit{Self Attention}), and ours (\textit{Text Guided Decoding}). In addition, we also implement the object query mechanism from MDETR~\cite{kamath2021mdetr} (\textit{Object Queries}) in our framework with the same training strategy (i.e., train MDETR for visual grounding from scratch, without multimodal pre-training). Table \ref{table:table_6} shows that our Dynamic MDETR outperforms other decoding strategies on both accuracy and efficiency. A possible reason is that using language as queries in our text guided decoding injects strong semantic information about the target object for accurate localization. 

\textbf{Comparisons with other dynamic visual grounding transformers.}
Existing dynamic transformers for reducing spatial redundancy are usually encoder-only architectures~\cite{rao2021dynamicvit, pan2021ia}. To verify the effectiveness of our proposed dynamic multimodal transformer decoder, we apply our proposed text guided decoding to TransVG and modify TransVG into a encoder-decoder structure. In addtion, we design two sampling methods based on the encoder-only baseline of TransVG~\cite{deng2021transvg}, including random sampling and attention based sampling. Specifically, we randomly sample 36 visual tokens from the visual feature sequence for random sampling, and sample 36 visual tokens with the highest attention score given by the [REG] token for attention based sampling. For fair comparison, we do sampling after 3 transformer encoder layers to keep the same encoder layers with our Dynamic MDETR. We also implement progressively reducing the number of sampled visual features for both TransVG and Dynamic MDETR. For TransVG, we conduct token sampling with a ratio of 60\% for the last three transformer encoder layers. For Dynamic MDETR, we sample 100, 64, and 36 points for the three layers of decoder respectively.

Table \ref{table:table_7} shows that encoder-decoder based methods achieve higher accuracies than all these encoder-only methods including TransVG and its dynamic variants, which demonstrates the effectiveness of encoder-decoder design. And our Dynamic MDETR equipped with both 2D sampling and text guided decoding gets the highest accuracy but with almost the lowest computational cost, proving the effectiveness and efficiency of our dynamic multimodal decoder. We also compare the latency running on a single Titan Xp GPU. Our Dynamic MDETR achieves evident inference acceleration compared to orignial TransVG. Dynamic MDETR decouples the visual grounding process into encoding and decoding phrases. Via this powerful and efficient decoding strategy, our model can adaptively sample a small number of informative visual features and is more effective than the other sampling strategies. Compared with 1D sampling, we analyze that our 2D adaptive sampling generates flexible sampling points in 2D spatial space based on the x-axis and y-axis offsets relative to the reference point, which helps to learn positional and geometric information effectively, thus leading to better localization performance.

\begin{table*}[t]
\caption{Comparison with state-of-the-art methods on RefCOCO/+/g. $^\ast$ denotes Dynamic MDETR with 6 encoder layers and 3 decoder layers. The default settings are 3 encoder layers and 3 decoder layers. $^\circ$ denotes Dynamic MDETR using additional training tricks including auxiliary losses~\cite{carion2020end} and freezing backbones at the beginning of training. Models marked in gray perform large-scale multimodal pre-training along the settings of MDETR.}
\centering
\scalebox{0.92}{
\begin{tabular}{c|c|ccc|ccc|ccc}
\hline
\multicolumn{1}{c|}{\multirow{2}{*}{Methods}} & \multicolumn{1}{c|}{\multirow{2}{*}{Backbone}} & \multicolumn{3}{c|}{RefCOCO}               & \multicolumn{3}{c|}{RefCOCO+}              & \multicolumn{3}{c}{RefCOCOg} \\
\multicolumn{1}{c|}{}                         & \multicolumn{1}{c|}{}                          & val   & testA & \multicolumn{1}{c|}{testB} & val   & testA & \multicolumn{1}{c|}{testB} & val-g   & val-u   & test-u   \\ \hline
\textbf{\textit{Two-stage:}}                                     &                                                &       &       &                            &       &       &                            &         &         &          \\
\multicolumn{1}{c|}{CMN~\cite{hu2017modeling}}                      & \multicolumn{1}{c|}{VGG16}                     & -     & 71.03 & \multicolumn{1}{c|}{65.77} & -     & 54.32 & \multicolumn{1}{c|}{47.76} & 57.47   & -       & -        \\
\multicolumn{1}{c|}{VC~\cite{zhang2018grounding}}                       & \multicolumn{1}{c|}{VGG16}                     & -     & 73.33 & \multicolumn{1}{c|}{67.44} & -     & 58.40 & \multicolumn{1}{c|}{53.18} & 62.30   & -       & -        \\
\multicolumn{1}{c|}{ParalAttN~\cite{zhang2018grounding}}                  & \multicolumn{1}{c|}{VGG16}                & - & 75.31 & \multicolumn{1}{c|}{65.52} & - & 61.34 & \multicolumn{1}{c|}{50.86} & 58.03       & -   & -    \\
\multicolumn{1}{c|}{LGRANs~\cite{wang2019neighbourhood}}                   & \multicolumn{1}{c|}{VGG16}                     & -     & 76.60 & \multicolumn{1}{c|}{66.40} & -     & 64.00 & \multicolumn{1}{c|}{53.40} & 61.78   & -       & -        \\
\multicolumn{1}{c|}{MAttNet~\cite{yu2018mattnet}}                  & \multicolumn{1}{c|}{ResNet-101}                & 76.65 & 81.14 & \multicolumn{1}{c|}{69.99} & 65.33 & 71.62 & \multicolumn{1}{c|}{56.02} & -       & 66.58   & 67.27    \\
\multicolumn{1}{c|}{DGA~\cite{yang2019dynamic}}                  & \multicolumn{1}{c|}{ResNet-101}                & - & 78.42 & \multicolumn{1}{c|}{65.53} & - & 69.07 & \multicolumn{1}{c|}{51.99} & -       & -   & 63.28    \\
\multicolumn{1}{c|}{RvG-Tree~\cite{yu2018mattnet}}                  & \multicolumn{1}{c|}{ResNet-101}                & 75.06 & 78.61 & \multicolumn{1}{c|}{69.85} & 63.51 & 67.45 & \multicolumn{1}{c|}{56.66} & -       & 66.95   & 66.51    \\
\multicolumn{1}{c|}{CMRE~\cite{hong2019learning}}                  & \multicolumn{1}{c|}{ResNet-101}                & - & 82.53 & \multicolumn{1}{c|}{68.58} & - & 75.76 & \multicolumn{1}{c|}{57.27} & -       & -   & 67.38    \\
\multicolumn{1}{c|}{NMTree~\cite{liu2019learning}}                   & \multicolumn{1}{c|}{ResNet-101}                & 76.41 & 81.21 & \multicolumn{1}{c|}{70.09} & 66.46 & 72.02 & \multicolumn{1}{c|}{57.52} & 64.62   & 65.87   & 66.44    \\ \hline
\textbf{\textit{One-stage:}}                                     &                                                &       &       &                            &       &       &                            &         &         &          \\
\multicolumn{1}{c|}{SSG~\cite{chen2018real}}                     & \multicolumn{1}{c|}{DarkNet-53}                & - & 76.51 & \multicolumn{1}{c|}{67.50} & - & 62.14 & \multicolumn{1}{c|}{49.27} & 47.47   & 58.80   & -    \\
\multicolumn{1}{c|}{FAOA~\cite{yang2019fast}}                     & \multicolumn{1}{c|}{DarkNet-53}                & 72.54 & 74.35 & \multicolumn{1}{c|}{68.50} & 56.81 & 60.23 & \multicolumn{1}{c|}{49.60} & 56.12   & 61.33   & 60.36    \\
\multicolumn{1}{c|}{RCCF~\cite{liao2020real}}                     & \multicolumn{1}{c|}{DLA-34}                    & -     & 81.06 & \multicolumn{1}{c|}{71.85} & -     & 70.35 & \multicolumn{1}{c|}{56.32} & -       & -       & 65.73    \\
\multicolumn{1}{c|}{ReSC~\cite{yang2020improving}}                     & \multicolumn{1}{c|}{DarkNet-53}                & 77.63 & 80.45 & \multicolumn{1}{c|}{72.30} & 63.59 & 68.36 & \multicolumn{1}{c|}{56.81} & 63.12   & 67.30   & 67.20    \\ \hline
\textbf{\textit{Transformer-based:}}                                     &                                                &       &       &                            &       &       &                            &         &         &          \\
\multicolumn{1}{c|}{VGTR~\cite{du2021visual}}                  & \multicolumn{1}{c|}{ResNet-50}                 & 78.29 & 81.49 & \multicolumn{1}{c|}{72.38} & 63.29 & 70.01 & \multicolumn{1}{c|}{55.64} & 61.64   & 64.19   & 64.01    \\
\multicolumn{1}{c|}{VGTR~\cite{du2021visual}}                  & \multicolumn{1}{c|}{ResNet-101}                 & 79.20 & 82.32 & \multicolumn{1}{c|}{73.78} & 63.91 & 70.09 & \multicolumn{1}{c|}{56.51} & 62.28   & 65.73   & 67.23    \\
\multicolumn{1}{c|}{VLTVG~\cite{yang2022improving}}                  & \multicolumn{1}{c|}{ResNet-50}                 & 84.53 & 87.69 & \multicolumn{1}{c|}{79.22} & 73.60 & 78.37 & \multicolumn{1}{c|}{64.53} & 72.53   & 74.90   & 73.88    \\ 
\multicolumn{1}{c|}{VLTVG~\cite{yang2022improving}}                  & \multicolumn{1}{c|}{ResNet-101}                 & 84.77 & 87.24 & \multicolumn{1}{c|}{80.49} & 74.19 & 78.93 & \multicolumn{1}{c|}{\underline{65.17}} & \underline{72.98}   & \underline{76.04}  & 74.18    \\ 
\multicolumn{1}{c|}{QRNet~\cite{ye2022shifting}}                  & \multicolumn{1}{c|}{Swin-S}                 & 84.01 & 85.85 & \multicolumn{1}{c|}{\textbf{82.34}} & 72.94 & 76.17 & \multicolumn{1}{c|}{63.81} & 71.89   & 73.03  & 72.52    \\
\multicolumn{1}{c|}{TransVG++~\cite{deng2022transvg++}}                  & \multicolumn{1}{c|}{ViT-B}                 & \textbf{86.28} & \underline{88.37} & \multicolumn{1}{c|}{\underline{80.97}} & \textbf{75.39} & \underline{80.45} & \multicolumn{1}{c|}{\textbf{66.28}} & \textbf{73.86}   & \textbf{76.18}  & \textbf{76.30}    \\

\hline
\multicolumn{1}{c|}{TransVG (Baseline)~\cite{deng2021transvg}}                  & \multicolumn{1}{c|}{ResNet-50}                 & 80.32 & 82.67 & \multicolumn{1}{c|}{78.12} & 63.50 & 68.18 & \multicolumn{1}{c|}{55.63} & 66.56   & 67.66   & 67.44    \\

\multicolumn{1}{c|}{Dynamic MDETR (Ours)}     & \multicolumn{1}{c|}{ResNet-50}                 & 80.47 & 82.63 & \multicolumn{1}{c|}{75.96} & 65.52 & 70.41 & \multicolumn{1}{c|}{57.68} & 66.87   & 69.20   & 68.56    \\
\multicolumn{1}{c|}{Dynamic MDETR$^\ast$ (Ours)}     & \multicolumn{1}{c|}{ResNet-50}                 & 81.62 & 83.85 & \multicolumn{1}{c|}{76.24} & 67.00 & 70.95 & \multicolumn{1}{c|}{58.13} & 68.04   & 70.14   & 69.57   \\
\multicolumn{1}{c|}{Dynamic MDETR$^\circ$ (Ours)}     & \multicolumn{1}{c|}{ResNet-50}                 & 82.80 & 84.99 & \multicolumn{1}{c|}{77.74} & 69.05 & 74.20 & \multicolumn{1}{c|}{60.17} & 69.43   & 70.73   & 70.68    \\

\multicolumn{1}{c|}{TransVG (Baseline)~\cite{deng2021transvg}}                  & \multicolumn{1}{c|}{ResNet-101}                & 81.02 & 82.72 & \multicolumn{1}{c|}{78.35} & 64.82 & 70.70 & \multicolumn{1}{c|}{56.94} & 67.02   & 68.67   & 67.73    \\ 
\multicolumn{1}{c|}{TransVG (Baseline)~\cite{deng2021transvg}}                  & \multicolumn{1}{c|}{CLIP-B}                 & 83.86 & 87.62 & \multicolumn{1}{c|}{79.32} & 70.50 & 77.18 & \multicolumn{1}{c|}{59.21} & 70.40   & 72.32   & 73.84    \\ 
\multicolumn{1}{c|}{Dynamic MDETR (Ours)}     & \multicolumn{1}{c|}{CLIP-B}                      & \underline{85.97} & \textbf{88.82} & \multicolumn{1}{c|}{80.12} & \underline{74.83} & \textbf{81.70} & \multicolumn{1}{c|}{63.44} & 72.21   & 74.14   & \underline{74.49}    \\ \hline
\rowcolor[HTML]{E7E6E6} 
\multicolumn{1}{c|}{MDETR~\cite{kamath2021mdetr}}                  & \multicolumn{1}{c|}{ResNet-101}                & 86.75 & 89.58 & \multicolumn{1}{c|}{81.41} & 79.52 & 84.09 & \multicolumn{1}{c|}{70.62} & -   & 81.64   & 80.89    \\ 
\rowcolor[HTML]{E7E6E6} 
\multicolumn{1}{c|}{UniTAB~\cite{yang2022unitab}}                  & \multicolumn{1}{c|}{ResNet-101}                & 88.59 & 91.06 & \multicolumn{1}{c|}{83.75} & 80.97 & 85.36 & \multicolumn{1}{c|}{71.55} & -   & 84.58   &  84.70   \\ 
\rowcolor[HTML]{E7E6E6} 
\multicolumn{1}{c|}{OFA~\cite{wang2022ofa}}                  & \multicolumn{1}{c|}{ResNet-101}                & 88.48 & 90.67 & \multicolumn{1}{c|}{83.30} & 81.39 & 87.15 & \multicolumn{1}{c|}{74.29} & -   & 88.07  & 88.78    \\  \hline

\end{tabular}
\label{table:table_8}
}
\end{table*}

\begin{table}[t]
\centering
\caption{Comparison with state-of-the-art methods on the test set of ReferItGame and Flickr30K Entities.}
\scalebox{0.9}{
\begin{tabular}{c|c|c|c}
\hline
\multirow{2}{*}{Methods} & \multirow{2}{*}{Backbone} & ReferItGame & Flickr30K \\
                         &                           & test        & test      \\ \hline
\textbf{\textit{Two-stage:}}               &                           &             &           \\
CMN~\cite{hu2017modeling}                      & VGG16                     & 28.33       & -         \\
VC~\cite{zhang2018grounding}                       & VGG16                     & 31.13       & -         \\
MAttNet~\cite{yu2018mattnet}                  & ResNet-101                & 29.04       & -         \\ 
Similarity Net~\cite{wang2018learning}                  & ResNet-101                & 34.54       & 60.89         \\
CITE~\cite{plummer2018conditional}                  & ResNet-101                & 35.07       & 61.33         \\
PRTC~\cite{kovvuri2018pirc}                  & ResNet-101                & 59.13       & 72.83         \\
DDPN~\cite{yu2018rethinking}                  & ResNet-101                & 63.00       & 73.30         \\
LCMCG~\cite{liu2020learning}                  & ResNet-101                & -       & 76.74         \\ \hline
\textbf{\textit{One-stage:}}               &                           &             &           \\
SSV~\cite{chen2018real}                     & DarkNet-53                & 54.24      & -     \\
FAOA~\cite{yang2019fast}                     & DarkNet-53                & 60.67       & 68.71     \\
RCCF~\cite{liao2020real}                     & DLA-34                    & 63.79       & -         \\
ReSC~\cite{yang2020improving}                     & DarkNet-53                & 64.60       & 69.28     \\\hline
\textbf{\textit{Transformer-based:}}    &       &       &             \\
VLTVG~\cite{yang2022improving}                 & ResNet-50                & 71.60      & 79.18 \\
VLTVG~\cite{yang2022improving}                  & ResNet-101                & 71.98       & 79.84 \\
QRNet~\cite{ye2022shifting}  & Swin-S   & \underline{74.61}       & 81.95 \\
TransVG++~\cite{deng2022transvg++}  & ViT-B                & \textbf{74.70}       & 81.49 \\

\hline
TransVG (Baseline)~\cite{deng2021transvg}                  & ResNet-50                 & 69.76       & 78.47     \\
Dynamic MDETR (Ours)     & ResNet-50                 & 69.95       & 79.21     \\
Dynamic MDETR$^\ast$ (Ours)     & ResNet-50                 & 71.13       & 79.50     \\
Dynamic MDETR$^\circ$ (Ours)     & ResNet-50                 & 69.48       & 79.32     \\
TransVG (Baseline)~\cite{deng2021transvg}                  & ResNet-101                & 70.73       & 79.10 \\
TransVG (Baseline)~\cite{deng2021transvg}                  & CLIP-B               & 71.21    & \textbf{82.48} \\
Dynamic MDETR (Ours)     & CLIP-B                      & 70.37      & \underline{81.89}     \\ \hline
\rowcolor[HTML]{E7E6E6} 
MDETR~\cite{kamath2021mdetr}                  & ResNet-101                & -      & 83.80 \\
\rowcolor[HTML]{E7E6E6} 
UniTAB~\cite{yang2022unitab}                  & ResNet-101            & -    & 79.58 \\ \hline

\end{tabular}
}
\label{table:table_9}
\end{table}

\begin{table}[t]
\centering
\caption{Results on validation set of VQA 2.0.}
\begin{tabular}{c|c}
\hline
Method     & Accuracy \\ \hline
Up-Down~\cite{anderson2018bottom}    & 63.15    \\
TransVG~\cite{deng2021transvg}    & 61.93   \\ 
Dynamic MDETR (ours)    & \textbf{64.21}    \\ \hline
\end{tabular}
\label{table:table_10}
\end{table}

\subsection{Comparison with the State-of-the-Art Methods}
\label{sec_4_4}

After the ablation studies, we are ready to compare our Dynamic MDETR with the existing state-of-the-art methods on these five visual grounding benchmarks, including RefCOCO~\cite{yu2016modeling}, RefCOCO+~\cite{yu2016modeling}, RefCOCOg~\cite{mao2016generation}, ReferItGame~\cite{kazemzadeh2014referitgame} and Flickr30K Entities~\cite{plummer2015flickr30k}. Table \ref{table:table_8} reports the performance comparison on RefCOCO, RefCOCO+ and RefCOCOg, and Table \ref{table:table_9} reports the performance comparisons on ReferItGame and Flickr30K Entities. 

\subsubsection{Comparison with non-transformer based methods} 
We can group the methods into two-stage methods and one-stage methods, and the one-stage methods include multimodal transformer based methods and non multimodal transformer based methods. We first compare with these two-stage methods and non multimodal transformer based one-stage methods. Table \ref{table:table_8} shows that our Dynamic MDETR (ResNet-50) with the default setting (3 encoder layers and 3 decoder layers) almost achieves better performance than all these two stage and non multimodal transformer based methods. Specifically, Dynamic MDETR (ResNet-50) outperforms ReSC~\cite{yang2020improving} by 2.84\% on val set of RefCOCO, 1.93\% on val set of RefCOCO+ and 3.75\% on test-u split of RefCOCOg. With the same number of encoder layers as TransVG, Dynamic MDETR$^\ast$ (ResNet-50) further improves its performance and outperforms these methods with a large margin, such as 3.99\% on val set of RefCOCO, 3.41\% on testA set of RefCOCO+ and 4.76\% on test-u split of RefCOCOg over ReSC~\cite{yang2020improving}. 

Table \ref{table:table_9} reports the performance comparison on ReferitGame and Flickr30K Entities, showing that our Dynamic MDETR outperforms all non multimodal transformer based methods. Remarkably, our Dynamic MDETR (ResNet-50) achieves 5.35\% (6.53\% for Dynamic MDETR$^\ast$) absolute improvement on test set of ReferitGame and 9.93\% (10.22\% for Dynamic MDETR$^\ast$) absolute improvement on test set of Flickr30K Entities, compared with ReSC~\cite{yang2020improving}. This superior performance demonstrates the effectiveness of our one-stage dynamic decoding paradigm for the visual grounding task.

\begin{figure*}[t]
    \centering
    \includegraphics[width=0.85\linewidth]{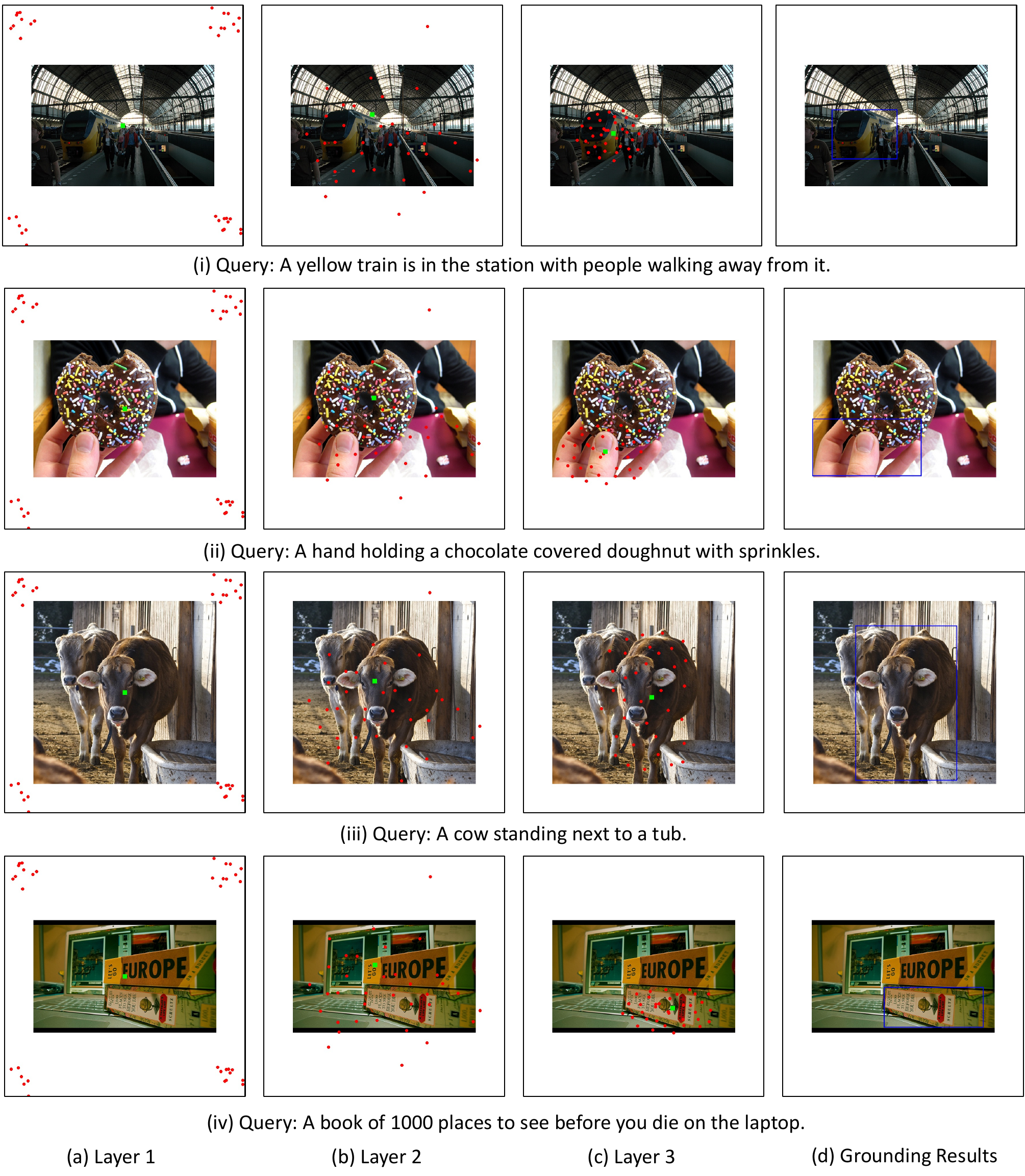}
    \caption{Sampled points in different decoder layers of Dynamic MDETR on RefCOCOg-umd validation set. The green point is the reference point, red points are sampled points and the blue box is the predicted bounding box.}
    \label{fig:fig_8}
\end{figure*}

\begin{figure*}[t]
    \centering
    \includegraphics[width=0.85\linewidth]{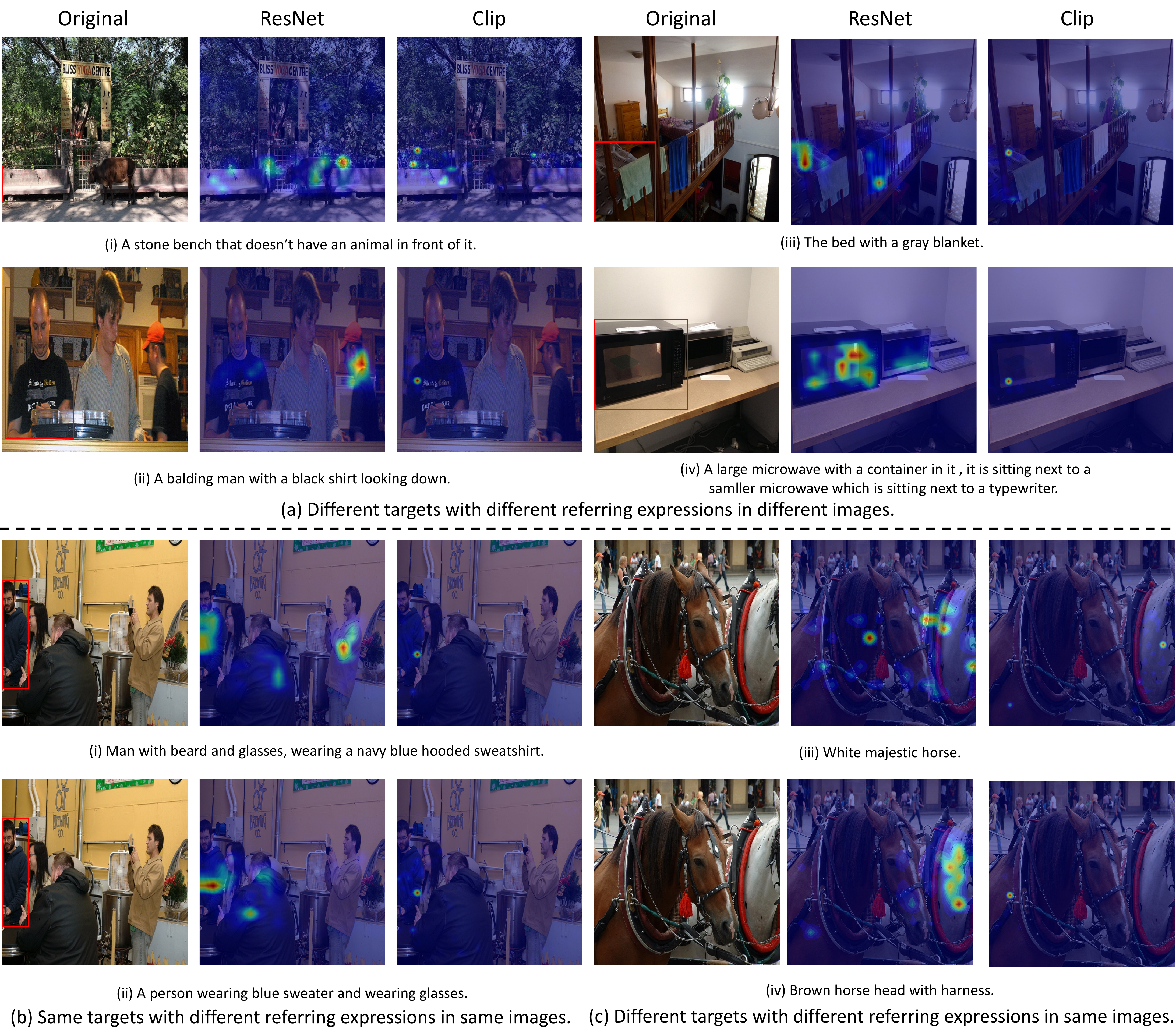}
    \caption{The qualitative results of Dynamic MDETR with ResNet and Clip as backbones on RefCOCOg-umd validation set. We visualize the averaged attention map of text tokens over the visual feature maps in the last encoder layer and show the target bounding box in red. Compared with ResNet, attentions with Clip are more sparse and more focused on the target object. Moreover, Dynamic MDETR with Clip is more robust to the diversity of language descriptions, and can localize the same target object given different referring expressions about the same targets.}
    \label{fig:fig_9}
\end{figure*}

\subsubsection{Comparison with transformer based methods} 

We also compare our Dynamic MDETR with other transformer based visual grounding methods. Table \ref{table:table_8} and Table \ref{table:table_9} show that Dynamic MDETR (ResNet-50) with a lighter encoder-decoder head slightly outperforms TransVG baseline of using the same backbone on all datasets except for RefCOCO. Specifically, we find our proposed Dynamic MDETR achieves an evident performance gain on two more challenging datasets, such as 2.23\% improvement on testA set of RefCOCO+ and 1.54\% improvement on val-u split of RefCOCOg. Surprisingly, Dynamic MDETR with ResNet-50 even performs better than TransVG with ResNet-101 on RefCOCO+ and RefCOCOg. With the same number of encoder layers, Dynamic MDETR$^\ast$ (ResNet-50) outperforms TransVG (ResNet-101) with a large margin at a marginal extra computational cost. There are a large number of location words in the referring expressions of RefCOCO, which are helpful for localization and reducing the difficulty of visual grounding. However, the expressions of RefCOCO+ are purely appearance based ones without any location words, and the expressions of RefCOCOg are longer and more complex. The performance improvements on these two challenging datasets show the strong multimodal reasoning ability and localization ability of Dynamic MDETR due to our adaptive modeling scheme.

\subsubsection{Transferring to the CLIP backbone}
Our proposed dynamic multimodal transformer decoder is a general visual grounding head.
We further investigate its scalability and generalization power by using the large-scale pre-trained CLIP backbone~\cite{radford2021learning}. To the best of our knowledge, we are the first work to adapt the CLIP to the task of visual grounding. 
CLIP is a two-branch architecture composed of a visual encoder and a text encoder, which is pre-trained on the large-scale image-text pair dataset. Specifically, we simply use its image encoder and text encoder without using any other neck module and directly place our dynamic multimodal decoder head on top. Our Dynamic MDETR (CLIP) uses a visual grounding head of 3-layer encoder and 3-layer decoder just like the ResNet-50 backbone.
As shown in Table \ref{table:table_8} and Table \ref{table:table_9}, Dynamic MDETR with CLIP as backbone outperforms its ResNet~\cite{he2016deep} counterpart with a large margin such as 11.29 points on RefCOCO+ testA set. Dynamic MDETR also gets higher accuracy than TransVG using CLIP as backbone, but with fewer computations on RefCOCO/+/g datasets. Moreover, Dynamic MDETR achieves competitive results compared to previous state-of-the-art methods, e.g., highest accuracy on testA set of RefCOCO dataset, and highest accuracy on val set of RefCOCO+ except for TransVG++. These superior performances demonstrate that our Dynamic MDETR framework can greatly benefit from more powerful foundation models, which can provide robust multi-modal representation for bridging the modality gap between images and texts. This important property can allow our proposed dynamic multimodal transformer decoder to be a general and flexible module for visual grounding, that could be effectively integrated with other foundation models in the future.

Finally, we compare with some large-scale pre-trained models (MDETR~\cite{kamath2021mdetr}, UniTAB~\cite{yang2022unitab} and OFA~\cite{wang2022ofa}) on the grounding task. It is worth noting that these multimodal pre-trained models along the setting of MDETR outperform the specific models designed for visual grounding task by a large margin due to the extra large-scale multi-modal pre-training. Therefore, it is hard to directly and fairly compare our dynamic MDETR with these large-scale pre-trained models. In general, we observe a performance gap between our method and these large-scale pre-trained models. In the future, we could resort to these powerful multimodal pre-trained models to further improve the performance of our dynamic MDETR.

\subsection{Transferring to VQA Task}
\label{sec_4_5}
To verify the task generalization ability of our model, we conduct experiments on another representative vision-language task, visual question answering (VQA). Due to the simple encoder-decoder structure, we can transfer our Dynamic MDETR to VQA task easily, by adding a classification head for VQA task. Following ~\cite{anderson2018bottom}, we validate the performance in VQA 2.0 dataset~\cite{goyal2017making}, and use train set for training and validation set for inference. We also implement TransVG for comparisons. The results are shown in Table \ref{table:table_10}. Dynamic MDETR outperforms the baseline~\cite{anderson2018bottom} and TransVG, showing its potential on VQA task and proving its generalization ability to other vision-language domains.

\subsection{Visualization}
\label{sec_4_6}
After the detailed ablation studies and quantitative comparisons with previous methods, we provide some qualitative analysis of our Dynamic MDETR framework to illustrate its effectiveness in this subsection. Specifically, we conduct two types of visualization analysis: plotting the sampled points in the dynamic decoder, and plotting the attention maps of text to images in the multimodal transformer encoder. We hope these visualization results could provide more insights into our Dynamic MDETR.

\textbf{Sampled points in different decoder layers.}
We plot the sampled points by our dynamic multimodal transformer decoder at different layers in Figure \ref{fig:fig_8}. The qualitative results show that as the number of decoder layers increases, the sampled points (red points) become more focused on the target object (blue box). Interestingly, we find that the initial sampled points fall out of the image and are uniformly distributed at the four corners of the image. We represent the points out of bounds with the features at the border points. We find that initial sampling points at the four corners are reasonable since we have no specific information on the target object, which can appear anywhere in the image. Thus, sampling from four corners can cover as many locations as possible to locate the target objects faster. We also find that our Dynamic MDETR equipped with 2D adaptive sampling can capture the distinguished properties of objects like "yellow", and complex relationships between the target object and context objects, like ``holding", ``next to", and ``left side". Surprisingly, our model can localize the book correctly, when given the text written on the book as the referring expression. It implies that our model has the ability of text recognition from an image.

\textbf{Text-to-image encoder attention map visualization.}
From previous results, we find that Dynamic MDETR with CLIP is much better than its ResNet counterpart. We also show some visualizations of text-to-image attention in the last multimodal transformer encoder layer of Dynamic MDETR with these two backbones to analyze the advantage of CLIP in Figure \ref{fig:fig_9}. Specifically, we average the attention maps of text tokens on the visual feature map and show the averaged attention map in the original image. We analyze three common situations in the visual grounding task. In Figure \ref{fig:fig_9} (a), we give some examples from different images with different text descriptions. Compared with ResNet, Dynamic MDETR with CLIP can focus on the target object from multiple objects more accurately, while ResNet will confuse the target with other objects and attend to wrong objects. 
We analyze that this advantage might be due to large-scale text and language pre-training, which can effectively bridge the gap between two modalities. This well aligned multimodal representation enables the attention map to be more focused on the target object, and generate lower attention values (much higher than the background area) around the reference point. We think this property is very important for the subsequent reference point localization in our decoder.
In Figure \ref{fig:fig_9} (b), we show some examples of the same targets with different referring expressions in the same images. We can see that when the referring expressions vary, the CLIP backbone still refers to the same targets. On the contrary, the attention with ResNet varies a lot and even attend to another object. This comparison demonstrates Dynamic MDETR with CLIP is more robust to the diversity of language descriptions. 
In Figure \ref{fig:fig_9} (c), we show some examples of the different targets with different referring expressions in the same images. We can see that Dynamic MDETR with CLIP accurately distinguishes the white horse and brown horse, while its counterpart with ResNet makes wrong predictions. This shows that Dynamic MDETR with CLIP can better understand the referring expressions, distinguish different objects and select the target object accurately in the same images.
Overall, the CLIP backbone can contribute to more accurate attention maps from text to image, which could lead to a more powerful Dynamic MDETR framework.

\section{Conclusion And Future Work}
In this paper, we have addressed the issue of spatial redundancy in images and proposed the Dynamic MDETR framework for efficient visual grounding. The core design of Dynamic MDETR is a dynamic multimodal transformer decoder, which is composed of a 2D adaptive sampling module and a text guided decoding module. Our dynamic multimodal transformer decoder can selectively sample a small number of informative points to greatly speed up the subsequent text guided decoding process. We conduct extensive experiments to demonstrate that the proposed encoder-decoder grounding framework is more effective and efficient than the encoder-only counterparts on the five benchmarks. In addition, to verify its generalization ability and scale up our Dynamic MDETR, we build the first one-stage CLIP empowered visual grounding framework, and achieve the state-of-the-art performance on these benchmarks. In the future, we plan to build Dynamic MDETR with more powerful multimodal pre-trained models and also extend it to video domain for temporal grounding.

In the future, we could improve our dynamic MDETR in several aspects. Specifically, the number of sampled points in each decoder layer is equal in our current design. Intuitively, fewer points are required in deeper layers, and adaptively adjusting the number of sampled points might be more optimal. We implement Dynamic MDETR-P, which progressively reducing the number of sampled points in the decoder layers, as an initial attempt. The results demonstrate the possibility of this idea. We could make the model automatically learn the number of sampling points in the future work. Moreover, the multimodal transformer encoder in our framework still contains dense connection and computation in a static way. Designing a dynamic multimodal encoder will make it a pure dynamic framework, and enable stronger dynamic capacity and higher efficiency. We also leave this idea as the future work.

\ifCLASSOPTIONcompsoc
  \section*{Acknowledgments}
\else
  \section*{Acknowledgments}
\fi

 This work is supported by the National Key R$\&$D Program of China (No. 2022ZD0160900), the National Natural Science Foundation of China (No. 62076119, No. 61921006), the Fundamental Research Funds for the Central Universities (No. 020214380091), and Collaborative Innovation Center of Novel Software Technology and Industrialization.

% or


% \appendices

% % you can choose not to have a title for an appendix
% % if you want by leaving the argument blank
% \section{}

% % use section* for acknowledgment
% \ifCLASSOPTIONcompsoc
%   % The Computer Society usually uses the plural form
%   \section*{Acknowledgments}
% \else
%   % regular IEEE prefers the singular form
% \fi


\ifCLASSOPTIONcaptionsoff
  \newpage
\fi



%\bibliographystyle{IEEEtran}


% \end{thebibliography}

% \newpage
% \clearpage

\bibliographystyle{IEEEtran}
\bibliography{ref}


\begin{IEEEbiography}[{\includegraphics[width=1in,height=1.25in,clip,keepaspectratio]{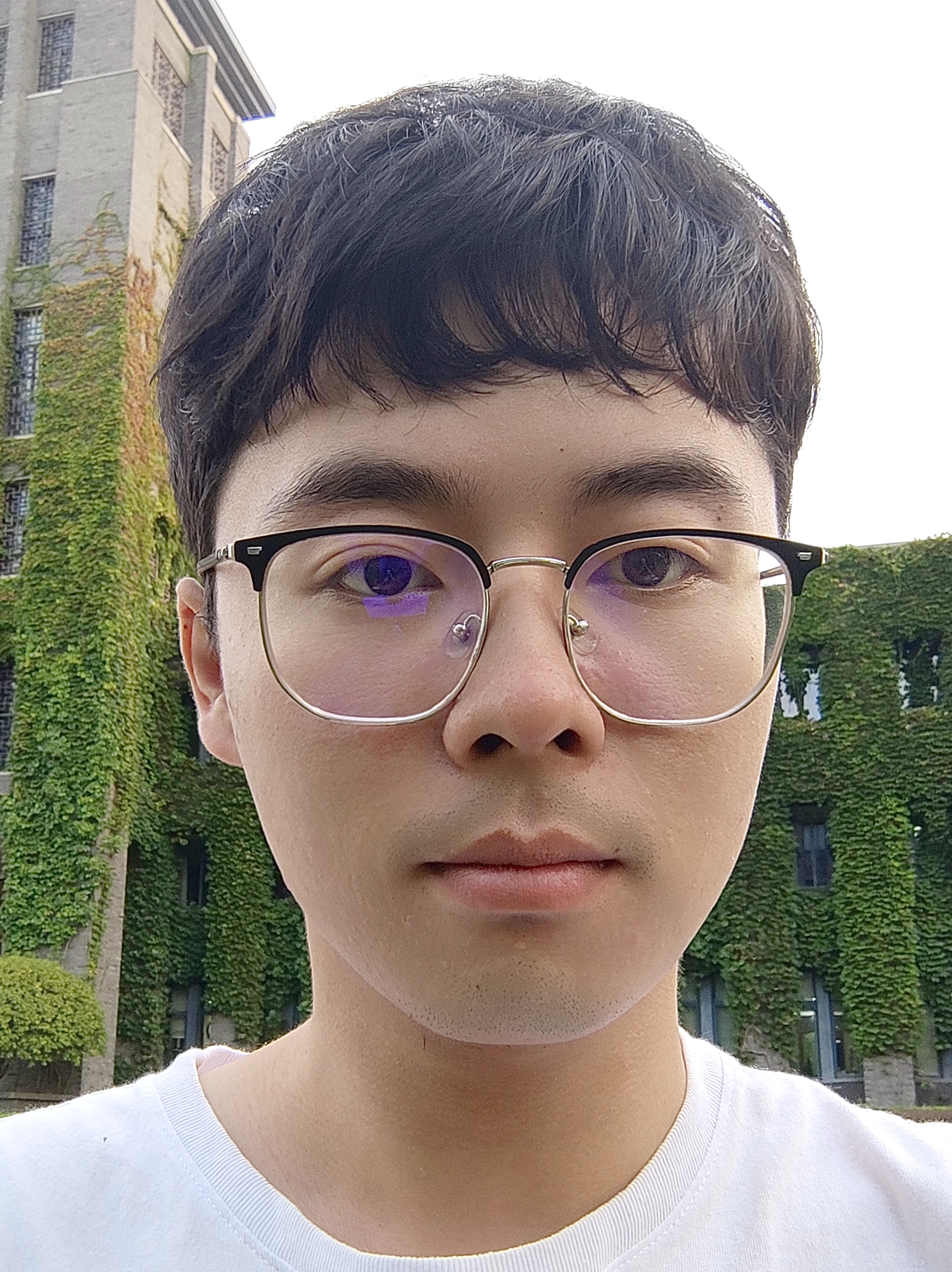}}]{Fengyuan Shi}
received the B.Sc. degree from  Department of Computer Science and Technology, Nanjing University, Nanjing, China, in 2021. He is currently a Ph.D. student in the Department of Computer Science and Technology, Nanjing University. His research interests include computer vision and deep learning.
\end{IEEEbiography}

\begin{IEEEbiography}[{\includegraphics[width=1in,height=1.25in,clip,keepaspectratio]{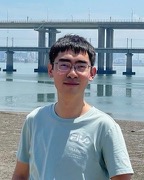}}]{Ruopeng Gao}
received the B.Sc. degree from  Department of Computer Science and Technology, Nanjing University, Nanjing, China, in June 2021. He is currently a Ph.D. student in the Department of Computer Science and Technology, Nanjing University, Nanjing, China. His research interests include computer vision and deep learning.
\end{IEEEbiography}


\begin{IEEEbiography}[{\includegraphics[width=1in,height=1.25in,clip,keepaspectratio]{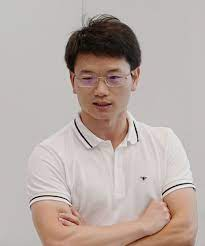}}]
{Weilin Huang} received the Ph.D. degree from The University of Manchester, U.K. He is currently a Senior Algorithm Expert with Alibaba Group, leading large-scale visual searching projects which connect billion-level users to billion-level products in real-world business applications. He worked as a Postdoctoral Researcher at the Visual Geometry Group (VGG), University of Oxford. He was an Assistant Professor with the Chinese Academy of Sciences. His research interests include computer vision, deep learning, and medical image analysis. He has served as a PC Member or a Reviewer for journals and conferences, including the IEEE TPAMI, IJCV, NeurIPS, ICLR, ICCV, CVPR, ECCV, ect. He is a member of the IEEE.
\end{IEEEbiography}

\begin{IEEEbiography}[{\includegraphics[width=1in,height=1.25in,clip,keepaspectratio]{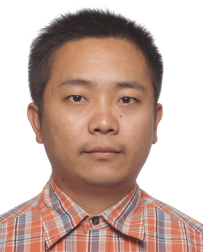}}]
 {Limin Wang} received the B.Sc. degree from Nanjing University, Nanjing, China, in 2011, and the Ph.D. degree from the Chinese University of Hong Kong, Hong Kong, in 2015. From 2015 to 2018, he was a Post-Doctoral Researcher with the Computer Vision Laboratory, ETH Zurich. He is currently a Professor with the Department of Computer Science and Technology, Nanjing University. His research interests include computer vision and deep learning. He was the first runner-up at the ImageNet Large Scale Visual Recognition Challenge 2015 in scene recognition, and the winner at the ActivityNet Large Scale Activity Recognition Challenge 2016 in video classification. He has served as a Area Chair for NeurIPS, CVPR, ICCV and is on the editorial board of IJCV. He is a member of the IEEE.
\end{IEEEbiography}


%\vfill


\end{document}